\documentclass{article}

\usepackage{amsmath,amsfonts,bm}

\def\eqref#1{equation~\ref{#1}}
\def\1{\bm{1}}

\def\rvb{{\mathbf{b}}}
\def\rvc{{\mathbf{c}}}

\def\rvs{{\mathbf{s}}}

\def\rvx{{\mathbf{x}}}
\def\rvy{{\mathbf{y}}}
\def\rvz{{\mathbf{z}}}

\def\rmG{{\mathbf{G}}}

\def\rmU{{\mathbf{U}}}
\def\rmV{{\mathbf{V}}}
\def\rmW{{\mathbf{W}}}

\DeclareMathAlphabet{\mathsfit}{\encodingdefault}{\sfdefault}{m}{sl}
\SetMathAlphabet{\mathsfit}{bold}{\encodingdefault}{\sfdefault}{bx}{n}

\def\gC{{\mathcal{C}}}

\def\gY{{\mathcal{Y}}}

\def\sR{{\mathbb{R}}}

\newcommand{\sigmoid}{\sigma}

\usepackage{microtype}
\usepackage{graphicx}
\usepackage{booktabs} % for professional tables
\usepackage{hyperref}

\usepackage{amsmath}
\usepackage{amssymb}
\usepackage{mathtools}
\usepackage{amsthm}
\usepackage[capitalize,noabbrev]{cleveref}
\usepackage{hyperref}
\usepackage{url}
\usepackage{float}
\usepackage{url}            % simple URL typesetting
\usepackage{booktabs}       % professional-quality tables
\usepackage{amsfonts}       % blackboard math symbols
\usepackage{nicefrac}       % compact symbols for 1/2, etc.
\usepackage{multirow}
\usepackage{dsfont}
\usepackage{soul}
\usepackage{adjustbox}
\usepackage{xspace}
\usepackage{bm}
\usepackage{amsmath}
\usepackage{setspace}                       % pleasant display of algorithm.
\usepackage{scrextend}
\usepackage{mathtools}
\usepackage{graphicx}
\usepackage{subcaption}
\usepackage{enumitem}
\usepackage{colortbl}
\usepackage{wrapfig}
\usepackage{lipsum}
\usepackage[ruled, noend, linesnumbered]{algorithm2e}

\usepackage[accepted]{icml2023}

\SetCommentSty{mycommfont}
\usepackage{xcolor,pifont}
\newcounter{asterfootnote}

\newcommand*\colourcheck[1]{%
  \expandafter\newcommand\csname #1check\endcsname{\textcolor{#1}{\ding{52}}}%
}

\colourcheck{blue}
\colourcheck{green}
\colourcheck{red}

\definecolor{pastelgreen}{rgb}{0.8, 0.866, 0.753}
\definecolor{pastelblue}{rgb}{0.788, 0.878, 0.925}
\definecolor{pastelorange}{rgb}{0.973, 0.894, 0.769}
\definecolor{pastelred}{rgb}{0.953, 0.835, 0.859}

\icmltitlerunning{Modality-Agnostic Variational Compression of Implicit Neural Representations}

\newcommand{\Lname}{Variational Compression of Implicit Neural Representations (VC-INR)\xspace} % large long name
\newcommand{\sname}{VC-INR\xspace} % short name

\begin{document}

\twocolumn[
\icmltitle{Modality-Agnostic Variational Compression of Implicit Neural Representations}



\icmlsetsymbol{equal}{*}

\begin{icmlauthorlist}
\icmlauthor{Jonathan Richard Schwarz}{equal,dm,ucl}
\icmlauthor{Jihoon Tack}{equal,kai}
\icmlauthor{Yee Whye Teh}{dm}
\icmlauthor{Jaeho Lee}{pos}
\icmlauthor{Jinwoo Shin}{kai}
\end{icmlauthorlist}

\icmlaffiliation{dm}{DeepMind}
\icmlaffiliation{ucl}{University College London}
\icmlaffiliation{kai}{KAIST}
\icmlaffiliation{pos}{POSTECH}

\icmlcorrespondingauthor{Jonathan Richard Schwarz}{schwarzjn@gmail.com}
%\icmlcorrespondingauthor{Firstname2 Lastname2}{first2.last2@www.uk}

\icmlkeywords{Data Compression, Neural data compression, Implicit Neural Representations, Neural Fields, Meta Learning}

\vskip 0.3in
]

\printAffiliationsAndNotice{\icmlEqualContribution} % otherwise use the standard text.

% ==========================================================
\begin{abstract}
We introduce a modality-agnostic neural compression algorithm based on a functional view of data and parameterised as an Implicit Neural Representation (INR). Bridging the gap between latent coding and sparsity, we obtain compact latent representations non-linearly mapped to a soft gating mechanism. This allows the specialisation of a shared INR network to each data item through subnetwork selection. After obtaining a dataset of such latent representations, we directly optimise the rate/distortion trade-off in a modality-agnostic space using neural compression. \Lname shows improved performance given the same representational capacity \textit{pre quantisation} while also outperforming previous quantisation schemes used for other INR techniques. Our experiments demonstrate strong results over a large set of diverse modalities using the same algorithm without any modality-specific inductive biases. We show results on images, climate data, 3D shapes and scenes as well as audio and video, introducing {\sname} as the first INR-based method to outperform codecs as well-known and diverse as JPEG 2000, MP3 and AVC/HEVC on their respective modalities.

\end{abstract}

% ==========================================================
\section{Introduction}
\label{sec:intro}

Data compression has become a critical problem in the modern era, as vast amounts of data is added to and transmitted through computer networks \cite{clissa2022survey} at previously unimaginable rates. While momentous progress has been made compared to naive representations, custom compression techniques are still developed for each modality at hand, carefully introducing inductive biases into new algorithms. While being an undoubtedly successful approach, it has limited the transfer of algorithmic ideas between techniques designed for different forms of data. More importantly, in certain engineering or scientific problems, vast amounts of data may be collected for which no generally accepted compression technique may be available (e.g. the AR/VR domain \citep{yang2022introduction}, point clouds, remote sensing or climate data), inhibiting progress in such fields.

In this paper, we join a recent group of researchers \citep[e.g.][]{dupont2021coin, dupont2022coinpp, schwarz2022meta} in arguing for a paradigm shift: Making modality-agnosticism a key guiding principle, we advocate for a single algorithmic workbench on which methods applicable to any type of data represented by a coordinate and feature space are developed. This would allow research effort to be pooled and any jointly developed model or learning improvement to benefit multiple downstream compression applications at once.

A promising approach towards realising this idea is the use of Implicit Neural Representations (INRs) or Neural Fields \citep[e.g.][]{tancik20fourier, sitzmann20implicit}. An INR relies on a functional interpretation of data, specifically as a mapping from coordinates to features (e.g. $(x,y) \rightarrow (r,g,b)$ for images), which is parameterised a neural network. INRs offer various attractive properties, including upsampling to arbitrary resolution \cite{chen2021learning} or a pathway to new approaches to applications such as generative modeling or classification \cite{dupont2022data}. For our purpose, the most intriguing property of the INR approach is its inherent modality-agnosticism, as any data point can in theory be represented provided it is expressed as a coordinate to feature mapping and thus learnable. Consequently, a learned INR is simply an encoding of the data point within the weights of a neural network, the efficient storage of which has received much attention at a time of ever increasing model capacity. We can thus state the second guiding principle of the work at hand: \textit{Data- as model-compression}.

This second principle distinguishes our ideas from much of the existing work on Neural Compression \citep[e.g.][]{balle2017end, balle2018variational, cheng2020learned}, which directly encodes a given data point into a codespace, hence relying on carefully designed modality-specific encoding and decoding networks (often called analysis/synthesis transforms). Throughout the manuscript, we will highlight how we overcome this limitation while building on rather than replacing the work from this community.

Among the recent work on compression with INRs on the other hand, various ideas for the efficient storage of INRs have been explored. So far, proposed compression and quantisation algorithms are relatively simple (e.g. Uniform Quantisation) or rely on a separate per-signal optimisation process \cite{strumpler2022implicit}, hence significantly increasing runtime. In addition, much of this work relies on ideas borrowed from Meta-Learning \cite{finn2017model} to decrease encoding times, which opens up various questions about the best trade-off between compact parameterisation and (Meta-) Learning algorithms. Therefore, despite significant efforts, a substantial gap still exists between INR-based compression and the hand-designed compression methods for certain modalities (e.g. JPEG 2000 for images, MP3 for audio).

In this paper, we improve INR-based compression in a two-fold approach (i) We experiment with advanced conditioning techniques resulting in better signal-reconstruction \textit{pre-quantisation} (ii) We overcome limitations of previously used quantisation techniques and introduce a learned quantiser, allowing us to maintain significantly higher reconstruction quality at lower file sizes \textit{post-quantisation}. Both directions of investigation adhere to the guiding principles of modality-agnosticism and the view of data as model compression. This presentation is not accidental, as we can think of the two axes of investigation as orthogonal algorithmic considerations. Indeed, any improvement in (i) increases the upper bound of performance maintained in the quantisation and entropy coding steps in (ii), while any improvement in (ii) reduces the gap between upper bound and actually realised performance.

\textbf{Contributions:}

\begin{itemize}[topsep=0.0pt,itemsep=0.5pt,leftmargin=5.5mm]
    \item \emph{Improved conditioning}:~ We propose a middle ground between recent sparsity and latent coding approaches to compact representations. The proposed technique introduces a non-linear mapping from a latent codes to a low-rank \textit{soft gating} matrix per layer, selecting a sub-network to represent a data item in an underlying INR. This is shown to learn more efficiently and result in better reconstructions compared to previous approaches. Our interpretation and experimental analysis shines new lights onto related ideas explored in other contexts.
    \item \emph{Improved compression}:~ We introduce a learned compressor pre-trained on compact latent codes representing training data. As such latent codes may be extracted from any modality, our proposed compressor operates fully modality-agnostic while making use of the same algorithmic insights previously only applicable to specific modalities.
\end{itemize}

We verify \sname on various data modalities, including image, voxels, scene, climate, audio, and video datasets. Overall, our experimental results demonstrate strong results, consistently outperforming previous INR-based compression methods and improving on popular compression schemes such as MP3 on audio and AVC/HEVC on video clips. In particular, \sname achieves a new state-of-the-art results on modality-agnostic compression with INRs, improving the Peak Signal to Noise Ratio (PSNR) on the same bits-per-pixel (bpp) bit rate by 3.3 dB for CIFAR-10 \cite{krizhevsky2009learning}, by 2 dB on Kodak\footnote{\footnotesize \url{https://www.kaggle.com/datasets/sherylmehta/kodak-dataset}} (both images), 3.5 dB for ERA5 (climate data) \cite{hersbach2019era5} and 9.5 dB for Librispeech (audio) \cite{panayotov2015librispeech} respectively. In addition, we outperform MP3 on Librispeech by 5.6 dB and HEVC on Videos by 8.8 dB.

Throughout this paper, we express a given data point $\rvx$ as a set of coordinates $\rvc\in\gC$ and real-valued features $\rvy\in\gY$ and its corresponding INR representation as $\boldsymbol\phi\in\mathbb{R}^D$. Whenever appropriate, we distinguish between $N$ data points using superscripts, i.e., $\{(\rvx^{i}, \boldsymbol\phi^{i})\}_{i=1}^N$ and individual coordinate/feature pairs using subscripts, i.e. $\rvx^{i}:=\{(\rvc_{j},\rvy_{j})\}_{i=1}^{M}$.

% ==========================================================
\section{Related work}
\label{sec:related}

\textbf{INRs} are neural networks approximating the functional mapping from coordinate to feature space. INRs are effective methods for modeling complex continuous signals, such as as 2D images \citep{chen2021learning}, 3D scenes \citep{park2019deepsdf}, videos \citep{kim2022scalable}, and are even applicable for modeling discrete data, e.g. graphs \citep{grattarola2022generalised}. To this end, several architectures have been proposed to capture  high-frequency signal details, examples being sinusoidal activations \citep{sitzmann20implicit}, positional encodings \citep{mildenhall20nerf}, and Fourier features \citep{tancik20fourier}. In practice, INRs are often specialised to each data item by fine-tuning from a shared initialisation \citep{tancik21learned}, drastically cutting the number of optimisation iterations.

\begin{figure*}[t]
\centering
\begin{subfigure}{0.3\textwidth}
\centering
    \includegraphics[width=0.9\linewidth]{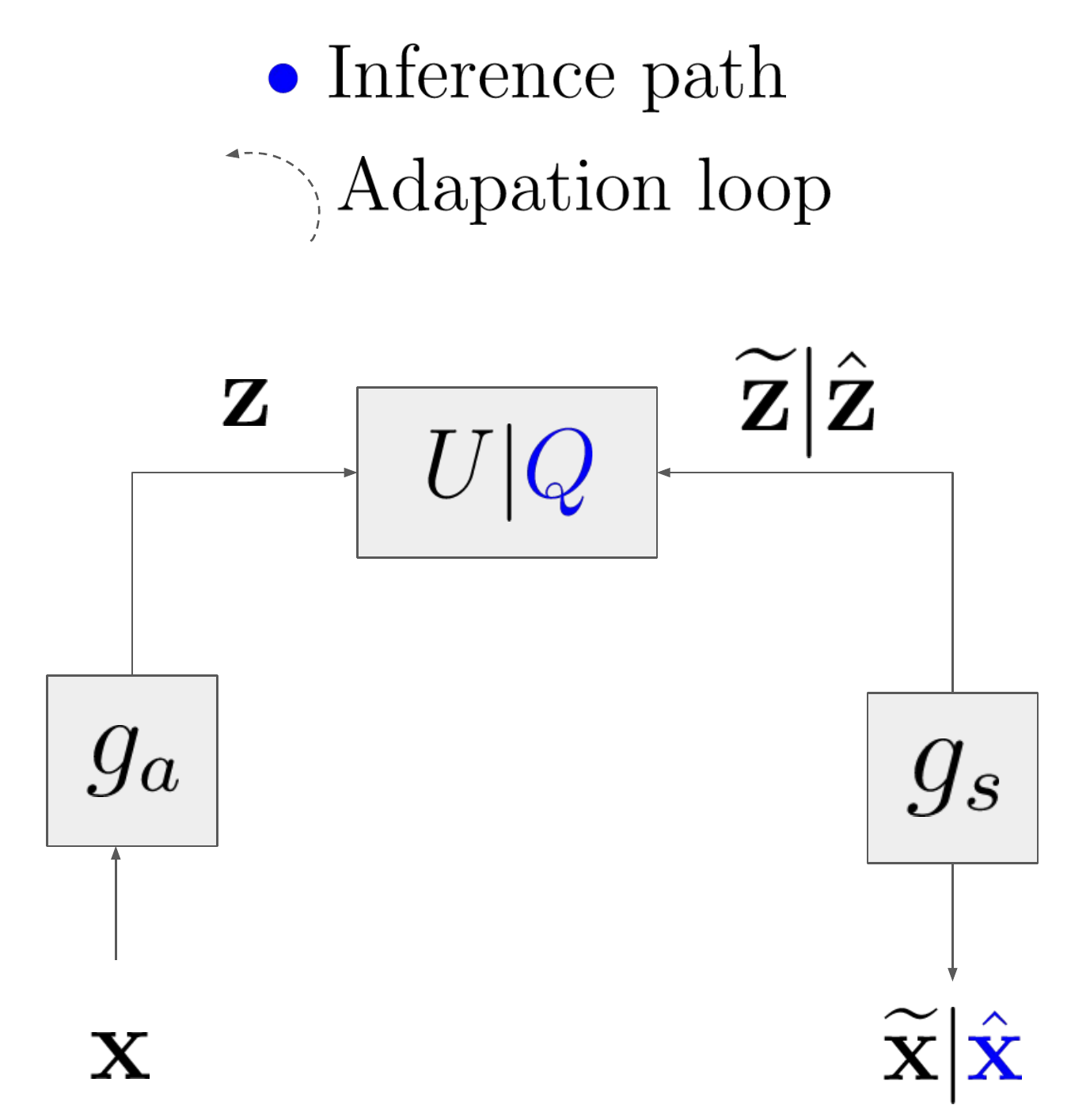}
    \caption{}
    \label{fig:operational_diagram_a}
\end{subfigure}%
\qquad
\begin{subfigure}{0.3\textwidth}
\centering
    \includegraphics[width=0.9\linewidth]{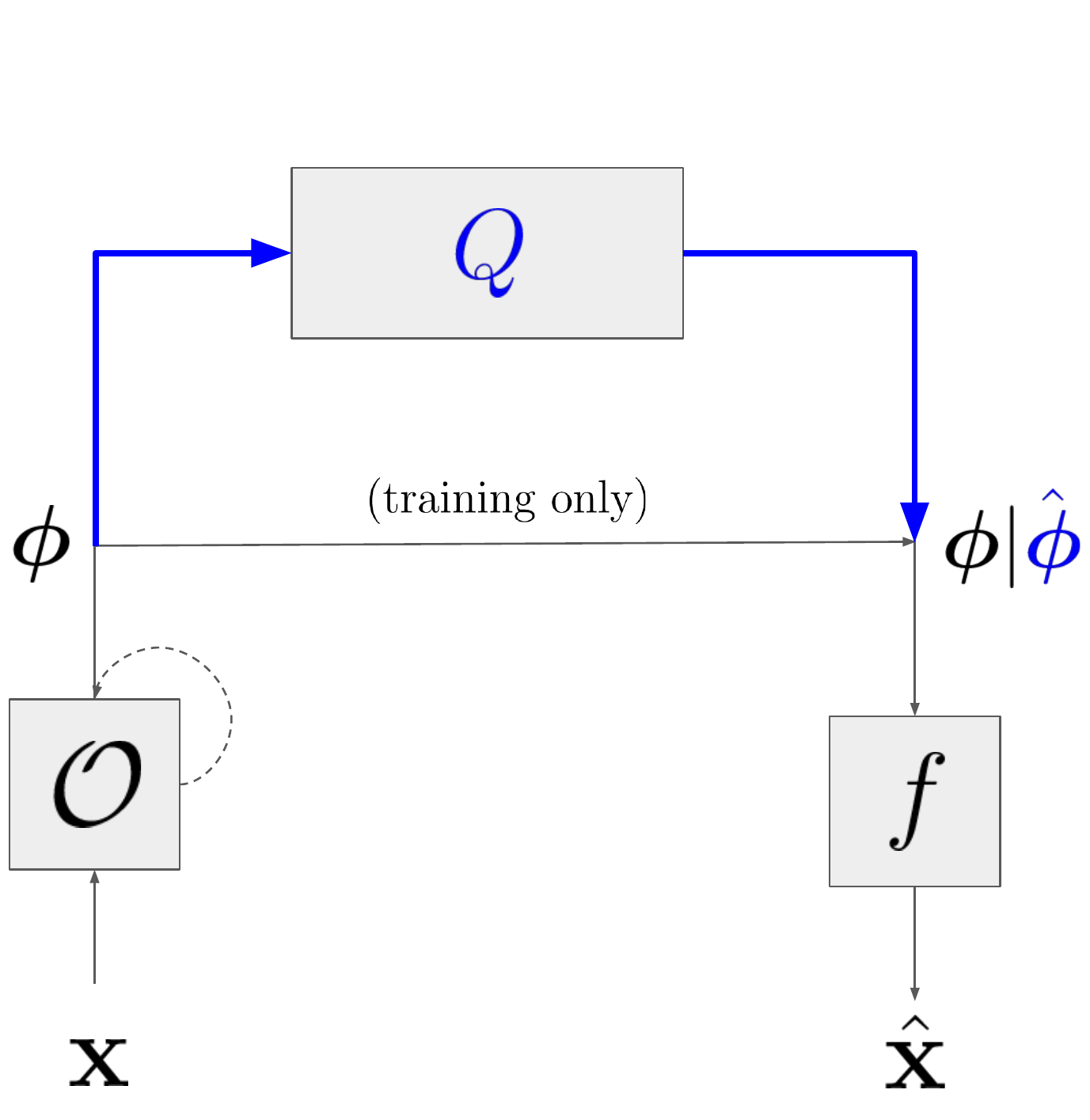}
    \caption{}
    \label{fig:operational_diagram_b}
\end{subfigure}
\qquad
\begin{subfigure}{0.3\textwidth}
\centering
    \includegraphics[width=0.9\linewidth]{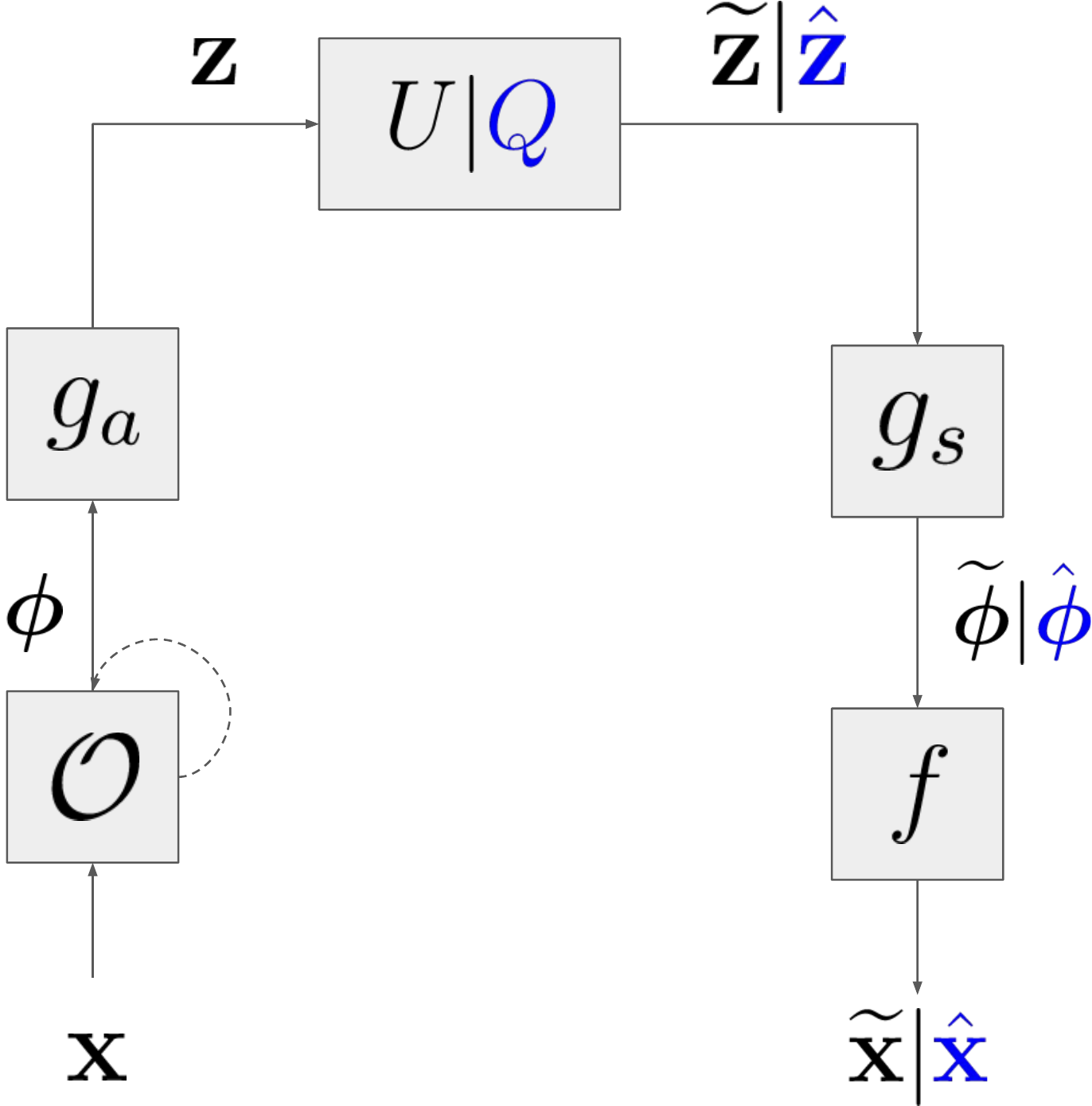}
    \caption{}
    \label{fig:operational_diagram_c}
\end{subfigure}
\caption{Operational diagrams of learned compression models. Inference time paths are shown in \textcolor{blue}{blue}. (a) Conventional neural compression \citep[e.g][]{balle2018variational} (b) Modality-agnostic neural compression with INRs \citep[e.g][]{dupont2022coinpp, schwarz2022meta} (c) Modality-agnostic variational compression of INRs is built upon the strengths of both techniques. $g_a, g_s$: Analysis/Synthesis; $f$: INR network; $U/Q$: Uniform noise/Quantisation; $\mathcal{O}$: Optimisation process; $\mathbf{x}, \boldsymbol{\phi}, \mathbf{x}$: data point, latent modulation, code element; $\widetilde{\mathbf{a}}, \hat{\mathbf{a}}$: Noisy version of $\mathbf{a}$ / Approximation of $\mathbf{a}$. For more details see text.}
\label{fig:operational_diagram}
\end{figure*}

\textbf{Neural compression} is an end-to-end autoencoder-based lossy compression framework aiming to directly minimise the inherent rate/distortion trade-off. This is based on a transform-coding approach \cite{goyal2001theoretical} shown in Figure \ref{fig:operational_diagram_a}, where a data item $\rvx$ is transformed into a latent code $\rvz$ through an analysis transform $g_a$. During training, quantisation is simulated through uniform noise ($\mathcal{U}$) resulting in a noisy $\widetilde{\rvz}$ and a corresponding reconstruction $\widetilde{\rvx} = g_s(\widetilde{\rvz})$ through the synthesis transform $g_s$. At test time, $\rvz$ is quantised (and entropy coded), resulting in codes and reconstructions $\hat{\rvz}, \hat{\rvx}$ respectively. Taking $g_a, g_s$ to be deep neural networks, the neural compression paradigm was introduced in \citep{balle2017end, theis2017lossy}, who make theoretical connections to variational inference. Recently, much of the recent work has focused on advanced designs of the entropy model, e.g. by using auto-regressive priors \cite{minnenbt18} or various forms of a hierarchical priors \citep{balle2018variational}, such as Gaussian mixture models (GMM) in \citep{minnen2018joint} and GMMs with attention modules \citep{cheng2020learned}. However, the majority of such neural compression techniques are typically focused on specific modalities, such as images \citep{lee2019context,agustsson2019generative, theis2022lossy} or videos \citep{lu2019dvc,habibian2019video,agustsson2020scale} and feature architectures specifically designed for such modalities, for instance convolutional architectures or the GDN activation function designed for natural images \cite{balle2015density}.

\textbf{Data compression with INRs} (Figure \ref{fig:operational_diagram_b}), introduced by \cite{dupont2021coin} as a modality agnostic compression method required long optimisation processes and architecture search to find a suitable rate/distortion trade-off. Following the wider INR literature \citep[e.g.][]{tancik21learned}, tabula-rasa learning was quickly replaced by a significantly faster Meta-Learning \cite{finn2017model} adaptation loop (shown as $\mathcal{O}$ in the diagram) while architecture search has been abandoned in favour of compact, instance specific representations $\boldsymbol\phi$ on which a deeper, shared INR $f$ is conditioned. The two mainstream approaches have been sparse representations \cite{lee2021meta, schwarz2022meta} implementing a close surrogate for the rate loss and/or FiLM-style modulations \cite{perez2018film, chan2021pi, mehta2021modulated} optionally linearly predicted from a compact latent code \cite{dupont2022data, dupont2022coinpp}.

Differing from the conventional neural compression workflow, methods following this paradigm either do not feature an explicit quantisation step \cite{dupont2021coin, lee2021meta} beyond default casting to 16-bit representation or rely on simple uniform quantisation based on first and second moment training statistics \citep{dupont2022coinpp, schwarz2022meta}. Recently, \citet{gordon2023quantizing} introduce an alternative quantisation scheme based on K-means clustering, avoiding the likely sub-optimal division of the quantisation space into equally sized regions. Crucially however, subsequent quantisation is not accounted for during training of the previous approaches, forgoing optimisation for deviations in the representations $\hat{\boldsymbol\phi}$. This is highlighted by a separate path at inference time in Figure \ref{fig:operational_diagram_b}. While advanced quantisation has been introduced \citep{strumpler2022implicit}, this requires additional training stages, thus increasing encoding runtime. \citet{damodaran2023rqat} is also similar to one aspect of this work by focusing on improving compression of INRs, showing strong improvements over COIN++ albeit only evaluating the method on images. In terms of applications of compression with INRs, \citet{huang2022compressing} show the large potential gains in climate applications while \citet{fons2022hypertime} focus on INRs for time series.

% ==========================================================
\section{Variational Compression of INRs}
\label{sec:method}
\subsection{Overview}
In contrast to the two approaches discussed in the previous section, we now present a computational framework which maintains modality-agnosticism while allowing the use of deep entropy coding. We show a high-level overview in Figure \ref{fig:operational_diagram_c}: The method can be best understood as an application of the non-linear transform coding paradigm (Figure \ref{fig:operational_diagram_a}) in the compact representation space of the INR approach (Figure \ref{fig:operational_diagram_b}).

More concretely, as in other INR techniques, we transform a data point $\rvx$ through an adaptation procedure $\mathcal{O}$ into a compact latent representation space $\boldsymbol\phi$. We can improve on the relatively simple quantisation techniques in prior works by employing non-linear transforming coding in $\boldsymbol\phi$ space, with its  analysis and synthesis transforms $g_a, g_s$ now operating on a modality-agnostic representation. This conceptionally simple change has the prime advantage of allowing the use of work from the neural compression literature with minimal changes (limited to simplification of $g_a$ \& $g_s$),  thus elevating conventional neural compression to a modality-agnostic paradigm. Compared to prior INR based compression, this allows the direction optimisation of the rate-distortion trade-off (as opposed to using a surrogate) using a deep entropy model. Moreover, a simple forward pass through $g_a$, subsequent quantisation $Q$ and then $g_s$ is preferable to an iterative technique such as quantisation aware training \citep{strumpler2022implicit} at inference time due to runtime considerations.

The rest of this section is split into the two axes of algorithmic improvements presented in this work: After giving a brief description of practical INR-learning on large datasets (Section \ref{sec:method-meta}), we then (i) Present an improved conditioning technique for specialising the shared base INR $f$ on the data-item specific representation $\boldsymbol\phi$ (Section \ref{sec:method-low-rank}) (ii) give a detailed discussion of the non-linear transform coding approach use (Section \ref{sec:method-variational}).

\begin{figure*}[t]
\centering
\begin{subfigure}{0.45\textwidth}
    \centering
    \includegraphics[width=0.8\linewidth]{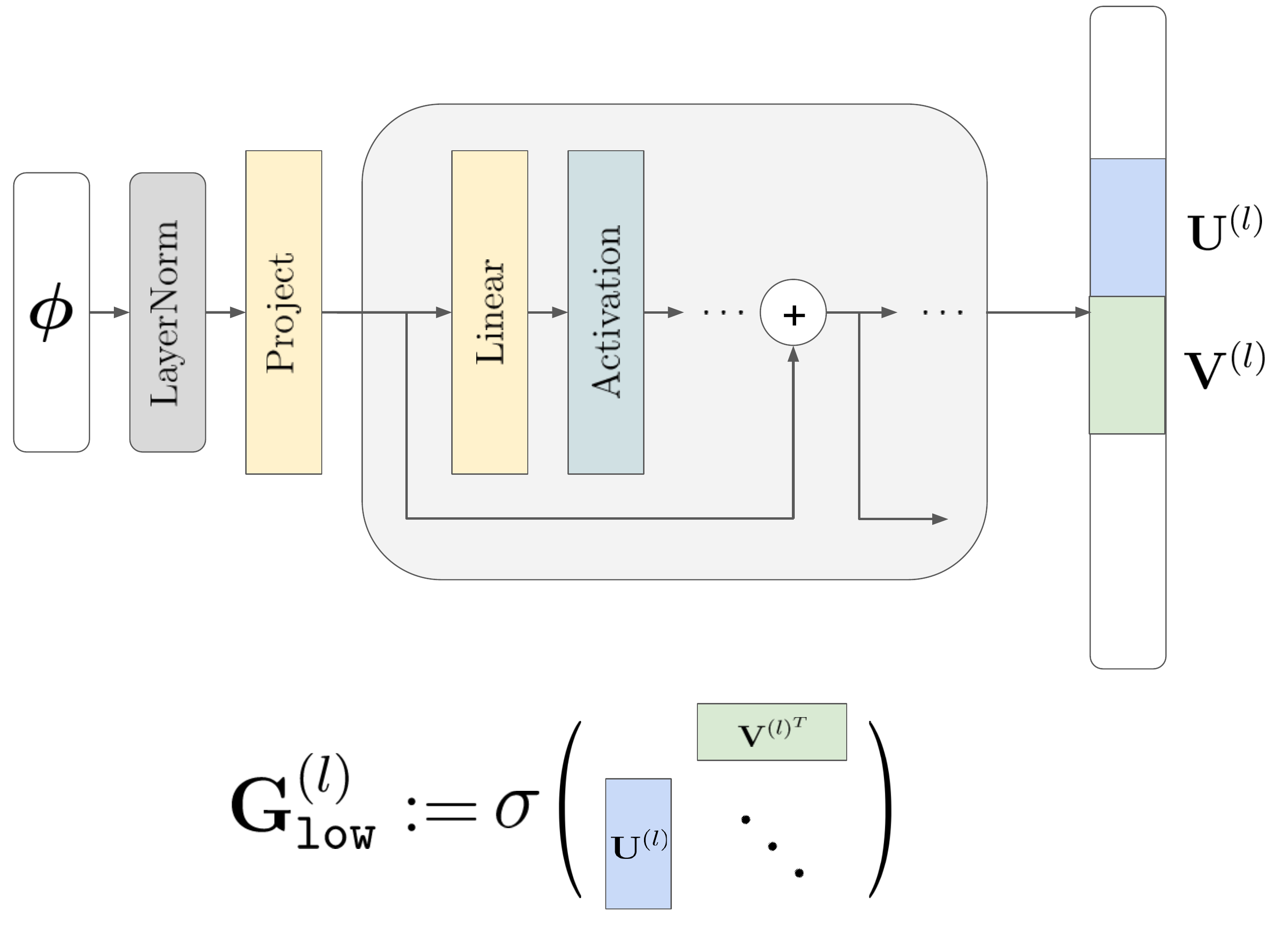}
    \caption{Non-linear projection from latent representation $\mathbf{\phi}$ to $\mathbf{G}_{\mathtt{low}}^{(l)}$.}
    \label{fig:low_rank}
\end{subfigure}%
\qquad
\begin{subfigure}{0.5\textwidth}
\centering
    \centering
    \includegraphics[width=0.8\linewidth]{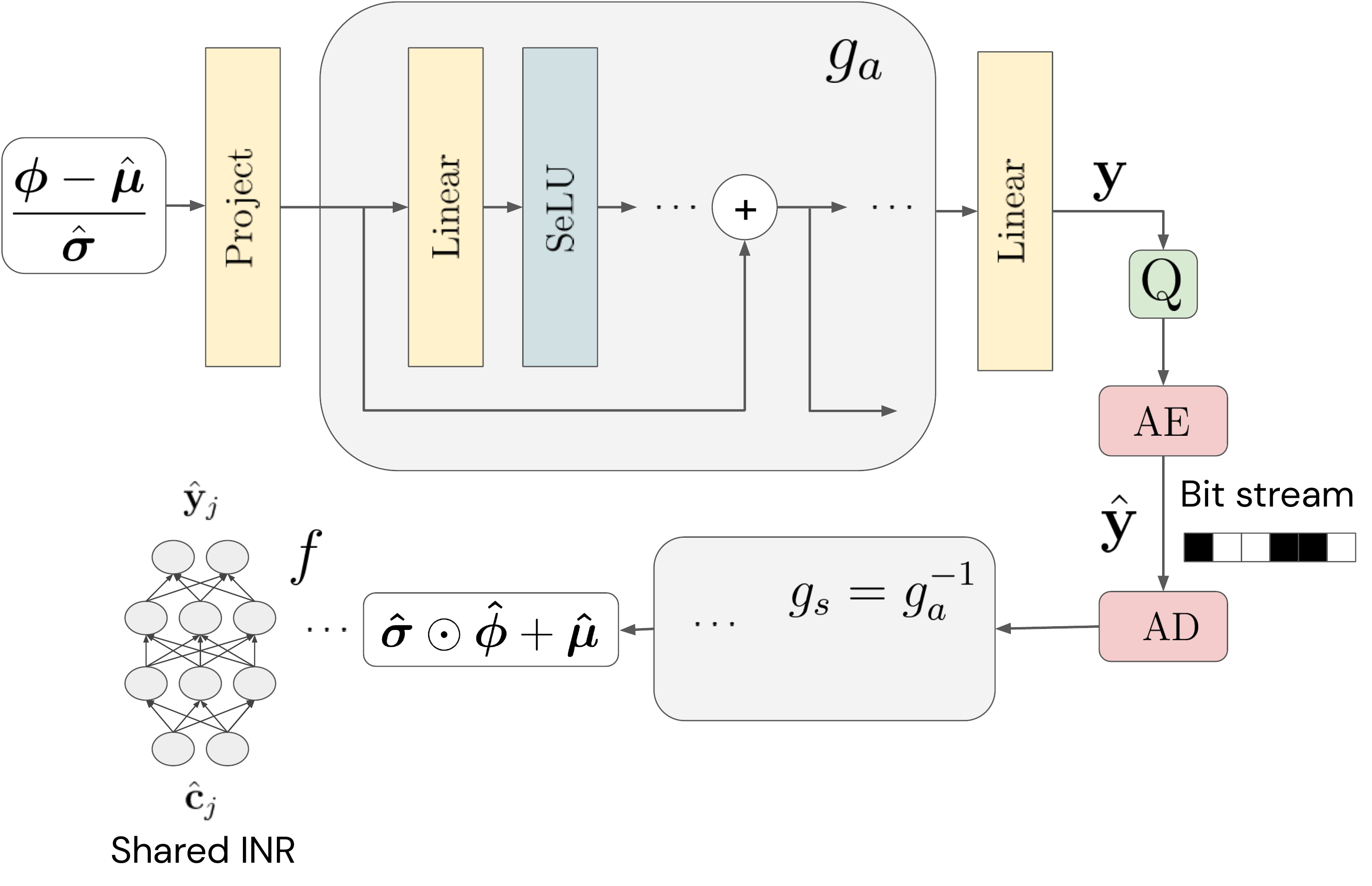}
    \caption{Non-linear transform coding in latent representation space.}
    \label{fig:tranform-coding}
\end{subfigure}
\caption{Architectural details of full model. AE/AD: Arithmetic Encoding/Decoding}
\label{fig:model_in_details}
\end{figure*}

\subsection{INR with data-wise modulations}
\label{sec:method-meta}

An INR is a function $f(\cdot;\boldsymbol\theta):\gC\to\gY$ representing a data point through a network with parameters $\boldsymbol\theta$. The INR objective is the mean-squared-error of predictions on the data point's coordinates $\{\rvc_{j}\}$ and the true features $\{\rvy_{j}\}$:
\vspace{-1mm}
\begin{equation}
    % \min_{\theta} \mathcal{L}_{\mathtt{INR}}(\theta,\rvx) =
    \min_{\boldsymbol\theta}\sum_{j=1}^{M}\big\lVert f(\rvc_{j};\boldsymbol\theta) - \rvy_{j}\big\rVert_{2}^{2}.
    \label{eq:mse-inr}
\end{equation}
\vspace{-4mm}

In practice, naive optimisation would require a large number of iterative steps and result in a set of high-dimensional parameter vectors $\{\boldsymbol\theta^i\}$ each representing a data point $\rvx^i$, making this an unattractive choice.

It is thus attractive to introduce a low dimensional data-item specific parameter $\boldsymbol\phi^i$ to model variations in $f$, while the much larger $\boldsymbol\theta$ is used to capture structure across a dataset. The shared INR $f(\cdot;\boldsymbol\theta)$ is specialised to $\rvx^i$ through $\boldsymbol\phi^i$ resulting in $f(\cdot;\boldsymbol\theta,\boldsymbol\phi^i)$. A reduction in the number of iterative steps per data item is achieved through Meta-Learning \citep{finn2017model}, allowing $\boldsymbol\phi^*$ for a test data point $\rvx^*$ to be obtained in a handful of optimisation steps (see Appendix for details on Meta-Learning).

Common ways to condition $f$ on $\boldsymbol\phi^i$ are layer-specific modulations $\rvs^{(l)}$ obtained by indexing into $\boldsymbol\phi^i$, i.e. $\boldsymbol\phi^i = [\rvs^{(1)}, \dots, \rvs^{(L)}]$ \citep{mehta2021modulated}. These modulations take the form of FiLM-style \citep{perez2018film} shifts, i.e. $\rvc^{(l-1)} \mapsto h(\rmW^{(l)} \rvc^{(l-1)} + \rvb^{(l)} + \rvs^{(l)})$, where $\rmW^{(l)},\rvb^{(l)}$ are shared weights and biases and $h$ is the activation function. To further reduce the size of $\boldsymbol\phi^i$, modulations of an $L$-layer INR $\rvs:=[\rvs^{(1)},\dots,\rvs^{(L)}]$ can be predicted from $\boldsymbol\phi^i$ using a shared linear mapping as $\rvs=\rmW^{'}\boldsymbol\phi+\rvb^{'}$ \citep{dupont2022data} or alternatively by pruning dimensions in $\boldsymbol\phi^i$ through sparsity \citep{schwarz2022meta}. Both techniques have their own drawbacks: Predictions of $\rvs$ from $\boldsymbol\phi$ have been challenging to train and so far been limited to linear mappings, thus lacking representational capacity. Sparsity techniques on the other hand require approximate inference, introducing additional complexity and various new hyperparameters.

\subsection{INR specialisation through subnetwork selection}
\label{sec:method-low-rank}

Instead, we take inspiration from both perspectives while overcoming their respective limitations. Following the sparsity paradigm, we observe that while a single network may be conditioned on potentially hundreds of distinct tasks through subnetwork selection \citep{frankle2018lottery, schwarz2021powerpropagation}, it is unclear whether this must be done through hard gating (i.e. requiring \textit{exact} zeros and ones) and thus require approximate inference. Indeed, recent work \citep{he2019task} suggests that soft-gating in the form of the output of a sigmoid $\sigmoid(x) = \frac{1}{1+e^{-x}}$ may be sufficient. In addition, it is clear that the idea of parametric predictions from $\boldsymbol\phi$ may in principle be used in conjunction with subnetwork selection, as a compact $\boldsymbol\phi$ could then concentrate its capacity on the non-sparse entries of $\mathbf{s}$, hence naturally combining both ideas.

To this end, we thus suggest the use of a non-linear prediction network mapping $\boldsymbol\phi$ to low-rank soft gating masks taking the same shape as the weight-matrices of each layer (see Figure \ref{fig:low_rank}). The functional form of the soft-gating masks is inspired by \citep{skorokhodov21adversarial} and takes the form of a low-rank matrix obtained through the outer product of two vectors non-linearly predicted from $\boldsymbol\phi$. This choice is sensible for two reasons: First, low-rank parameterisation is widely used as an effective tool for parameter reductions \citep{phan2020stable} and secondly, such modulation have shown potential in representing complex signals such as high-resolution images \citep{skorokhodov21adversarial} and videos \citep{yu2022generating}. Formally, given the activations of the preceding layer $\rvc^{(l-1)}$, the transformation of each layer $l$ is
\begin{align}
    \rvc^{(l-1)} \mapsto &\sin(\omega_{0} (\rmG_{\mathtt{low}}^{(l)} \odot \rmW^{(l)} ~ \rvc^{(l-1)} + \rvb^{(l)}))\\
    &\rmG_{\mathtt{low}}^{(l)}:=  \sigma(\rmU^{(l)}\rmV^{(l)\top}),
    \label{eq:siren-low-rank}
\end{align}
where $\rmU^{(l)},\rmV^{(l)}\in\sR^{m\times d}$ are data specific parameters with $d\ll m$, $\odot$ is element-wise multiplication and $\sigma(\cdot)$ the sigmoid operator. Here, we use sinusoidal activation function with its hyperparameters $\omega_0\in\mathbb{R}^{+}$ introduced for INRs in \citep{sitzmann20implicit}.
The \textbf{central hypothesis} of this approach is that $\rmG_{\mathtt{low}}^{(l)}$ acts as a subnetwork selection method, effectively determining and scaling the entries in each weight matrix $\rmW^{(l)}$ that allow accurate modeling of the data point at hand. We show evidence for this phenomenon in the experimental section.

As before, we can reduce the dimensions of low-rank modulation further, obtaining $[\rmU^{(1)},\rmV^{(1)},\cdots,\rmU^{(N)},\rmV^{(N)}]$ directly from the compact representation $\boldsymbol\phi$ by predicting a long vector, subsequently reshaped into the respective matrices. Unlike existing methods utilising a linear mapping \citep{dupont2022data,dupont2022coinpp}, we use deep residual networks to increase the expressive power, enabled through various stabilisation techniques:

\textbf{Stabilisation techniques} In line with prior work, we find the direct optimisation of non-linear networks via Meta-Learning to be unstable and under-performing. As low-rank parameterisations are also known to suffer from stability issues, the direct use yield unsatisfactory results. Instead, we suggest three stabilising techniques. (1) First, we propose the normalisation of the modulation $\boldsymbol\phi$ with LayerNorm \citep{ba2016layer}, i.e., $\boldsymbol\phi \mapsto \mathtt{LayerNorm(\boldsymbol\phi)}$ as in Fig \ref{fig:low_rank}. Intuitively, this results in higher order gradient optimisation becoming more stable as a normalisation scheme reduces the sharpness of the gradients \citep{santurkar2018does,xu2019understanding}. (2) We find residual connections and increasing layer widths (up to computational limits) to aid gradient propagation and significantly increase the performance of non-linear networks. (3) We hypothesise that the sigmoidal bounding of $\rmG_{\mathtt{low}}^{(l)}$ itself has a stabilising effect, preventing the matrix norm from divergence. % \citep{skorokhodov21adversarial}.

At this point it is worth noting that the combination of subnetwork selection techniques and non-linear predictors are not unique to compression and indeed may be beneficial in the wide array of downstream applications made possible through the INR paradigm \citep{dupont2022coinpp}. Next, we explain the subsequent quantisation of $\boldsymbol\phi$.

\subsection{Variational compression of modulations}
\label{sec:method-variational}

The key to using non-linear transform coding in a modality-agnostic paradigm is the observation that $\boldsymbol\phi$ may be obtained from data of any kind. For a given modulation $\boldsymbol\phi$, our goal is now to encode the modulation into a code $\rvz = g_a(\boldsymbol\phi)$ with low Shannon cross-entropy (its rate) and a reconstruction  $\hat{\boldsymbol\phi} = g_s(\hat{\rvz})$ with low distortion from $\boldsymbol\phi$ after quantisation $\hat{\rvz} = Q(\rvz) = \text{round}(\rvz)$. Because $\hat{\rvz}$ is discrete, it can be losslessy compressed using \textit{entropy coding} such as arithmetic or Huffman coding \citep{salomon2004data} to obtain a bit stream.

Here, we use the deep-factorised prior introduced for images in \citep{balle2017end} and used as the basis of many follow-up works. The authors establish the interpretation of a relaxed rate-distortion performance as variational autoencoder under a specific generative and inference model, lending the name \sname to our method.

We state the compression loss as the sum of (i) the \emph{rate} of the code and (ii) the \emph{distortion} of the recovered signal:

\begin{equation}
\begin{split}
    &\mathcal{L}_{\mathtt{compress}}(\boldsymbol\pi_{a},\boldsymbol\pi_{s}, \rvx, \boldsymbol\phi) \\
    &= \mathcal{L}_{\mathtt{rate}} + \lambda\mathcal{L}_{\mathtt{distortion}}\\
    &=-\log_{2}[p_{\hat{\rvz}}(Q(g_{a}(\boldsymbol\phi;\boldsymbol\pi_{a})))]
    + \lambda \mathcal{L}_{\mathtt{MSE}}(g_{s}(\hat{\rvz};\boldsymbol\pi_{s}), \boldsymbol\phi)
    \label{eq:rate-distor}
\end{split}
\end{equation}

with $p_{\hat{\rvz}}$ the entropy model, $\mathcal{L}_{\mathtt{MSE}}$ the mean squared error (MSE),
and  $\boldsymbol\pi_{a}, \boldsymbol\pi_{s}$ parameters of the analysis and synthesis transforms. The reconstruction $\hat{\boldsymbol\phi}$ is decoded from the quantised code $\hat{\rvz}$. To optimise this loss, we follow \cite{balle2017end} by approximating the discrete quantisation with uniform noise $\mathcal{U}(-\frac{1}{2},\frac{1}{2})$ to generate a noisy code $\tilde{\rvz}$ and use the differentiable prior $p_{\tilde{\rvz}}$ with a non-parametric piecewise linear density model \citep{balle2018variational}.

We show architectural details in Figure \ref{fig:tranform-coding}:  Differing from the typical design of $g_a, g_s$ we do not make use of activations with local gain control and find the SeLU activation \citep{klambauer2017self} sufficient. In addition, as the vectors $\boldsymbol\phi$ are flat regardless of modality, we can simplify the design of both networks to residual MLPs, removing another form of modality specificity. Finally, we note that the distortion term in Equation (\ref{eq:rate-distor}) is merely a surrogate for the real reconstruction quality of the data $\rvx$. We thus modify $\mathcal{L}_{\mathtt{distortion}}$ to measure distortion on data directly:
\begin{equation}
    \mathcal{L}_{\mathtt{distortion}} = \mathcal{L}_{\mathtt{MSE}}(f(\cdot; \boldsymbol\theta, g_{s}(\hat{\rvz};\boldsymbol\pi_{s})), \rvy)
\end{equation}
which we observe to result in the highest quality reconstructions. At this point we emphasise that advanced techniques \citep{balle2018variational, cheng2020image} may be straight-forwardly introduced.

% ==========================================================
\section{Experiments}
\label{sec:exp}

\begin{figure*}[t]
\centering
\begin{subfigure}{0.33\textwidth}
\centering
    \includegraphics[width=0.9\linewidth]{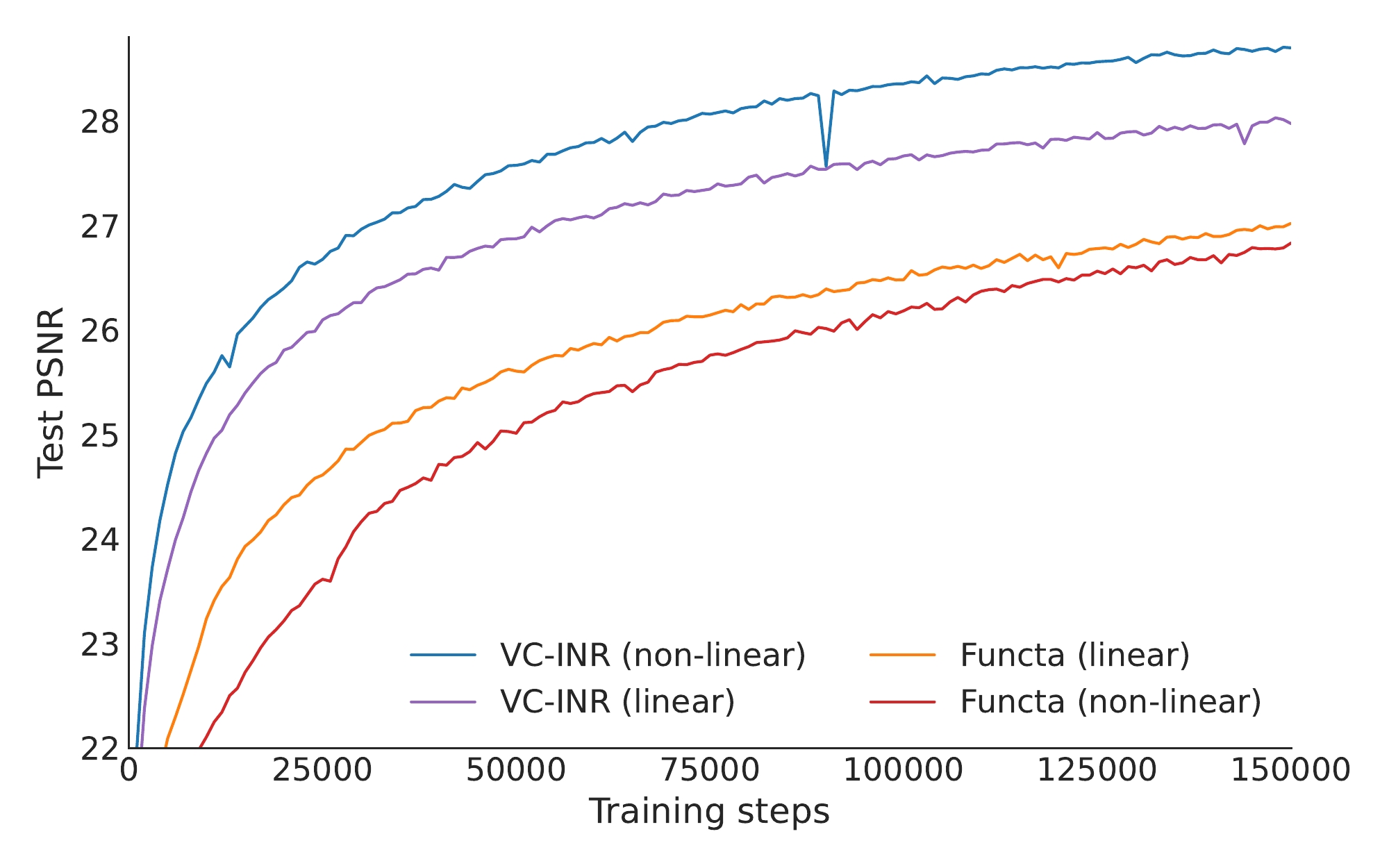}
    \caption{Learning curves}
    \label{fig:curves}
\end{subfigure}
\begin{subfigure}{0.33\textwidth}
\centering
    \includegraphics[width=0.9\linewidth]{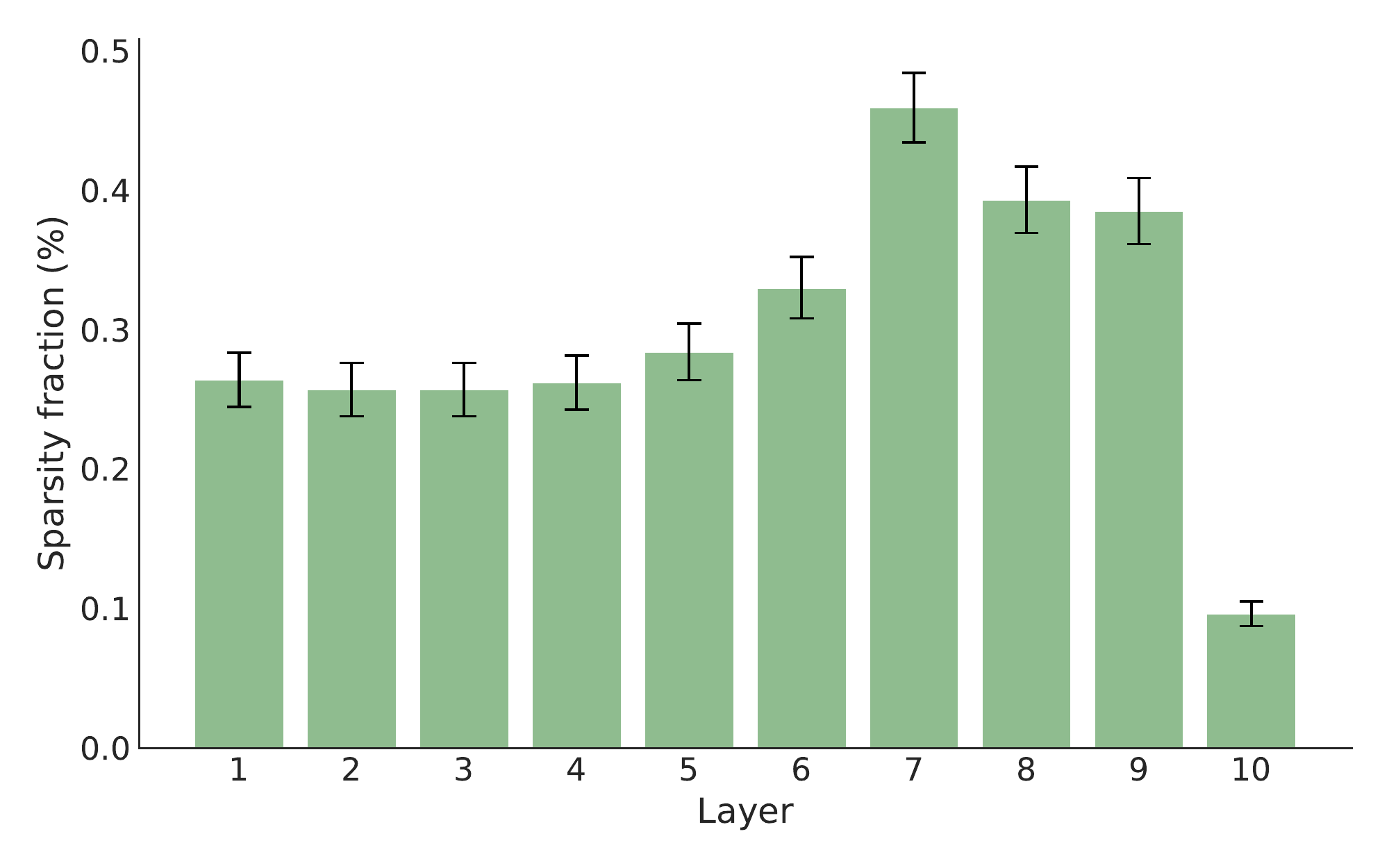}
    \caption{Sparsity pattern}
    \label{fig:sparsity}
\end{subfigure}%
\begin{subfigure}{0.33\textwidth}
\centering
    \includegraphics[width=0.75\linewidth]{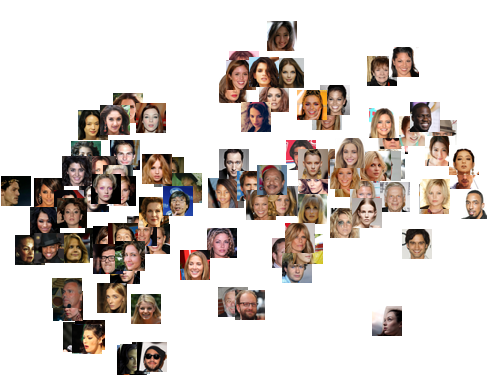}
    \caption{Mask clustering on CelebA-HQ}
    \label{fig:mask_celeba}
\end{subfigure}
\caption{Analysis of the VC-INR on CelebA-HQ (a) Learning curves of Functa, linear \& non-linear VC-INR models (b) Sparsity patterns of adapted weights throughout the network (c) t-SNE \citep{van2008visualizing} visualisation of masks $\rmG_{\mathtt{low}}$ after adaptation.}
\label{fig:modulation_analysis}
\end{figure*}

So far, we have discussed a two-fold approach: (i) Advanced conditioning to better capture an underlying signal within a fixed representation \textit{pre-quantisation}. (ii) Variational compression subsequently trained on datasets of such representations. In our empirical evaluation, we will first demonstrate the effectiveness of (i) in isolation (as its results is an upper bound for distortion performance). We then demonstrate the combination of both ideas on a range of compression problems. This will help clearly delineate performance gains as well as provide additional insights into the technique. Throughout the section, we primarily evaluate the performance using the Peak Signal to Noise Ratio (PSNR): $-10 \cdot \log_{10}$(MSE), where MSE is the mean-squared error between the original and the reconstructed signal.

\subsection{Effectiveness of advanced conditioning}
\label{exp:modulation}

\begin{table}[h]
\caption{Results for various latent modulation sizes. Shown is voxel accuracy (ShapeNet10) and PSNR (others).}
\vspace{1mm}
\label{tab:modulation}
\centering
\footnotesize
\setlength\tabcolsep{5pt}
\begin{tabular}{l|c|lllll@{}}
\toprule
\footnotesize
Dataset & Model &
\multicolumn{5}{c}{Performance @ $\dim(\boldsymbol\phi)$} \\ 
&                     & 64 & 128 & 256 & 512  & 1024  \\ \midrule
\rowcolor{pastelblue} ERA5 ($4\times$) & Functa & 43.2 & 43.7 & 43.8 & 44.0 & 44.1 \\
\rowcolor{pastelblue}             & MSCN   & 44.6 & 45.7 & 46.0 & 46.6 & 46.9 \\
%\rowcolor{pastelblue}                            & VC-INR (Linear) & 45 & 46.1 & 47.3 & 48.6 & 49.6 \\
\rowcolor{pastelblue}             & VC-INR & \textbf{45.0} & \textbf{46.2} & \textbf{47.6} & \textbf{49.0} & \textbf{50.0} \\
\midrule
\rowcolor{pastelorange}CelebA-HQ   & Functa & 21.6 & 23.5 & 25.6 & 28.0 & 30.7 \\
\rowcolor{pastelorange}           & MSCN   & 21.8 & 23.8 & 25.7 & 28.1 & \textbf{30.9} \\
%\rowcolor{pastelorange}                            & VC-INR (Linear) & 21.9 & 23.7 & 25.7 & 28.0 & 30.7 \\
\rowcolor{pastelorange}           & VC-INR & \textbf{22.0} & \textbf{23.9} & \textbf{26.0} & \textbf{28.3} & 30.8 \\

\midrule
\rowcolor{pastelred}SRN Cars      & Functa & 22.4 & 23.0 & 23.1 & 23.2 & 23.1 \\
\rowcolor{pastelred}              & MSCN & 22.8 & \textbf{24.0} & \textbf{24.3} & 24.5 & 24.8 \\   
%\rowcolor{pastelred}             & VC-INR (Linear) & 23.6 & 23.7 & 24.0 & 24.9 & 25.3 \\
\rowcolor{pastelred}              & VC-INR & \textbf{23.9} & \textbf{24.0} & \textbf{24.3} & \textbf{25.2} & \textbf{25.5} \\
\midrule
\rowcolor{pastelgreen}ShapeNet10 & Functa & $_{99}$.30 & $_{99}$.40 & $_{99}$.44 & $_{99}$.50 & $_{99}$.55 \\
\rowcolor{pastelgreen}            & MSCN & $_{99}$.43 & $_{99}$.50 & $_{99}$.56 & $_{99}$.63 & $_{99}$.69 \\ 
%\rowcolor{pastelgreen}            & VC-INR (Linear) & 99.47 & 99.52 & 99.58 & 99.63 & 99.7 \\
\rowcolor{pastelgreen}            & VC-INR & $_{99}$\textbf{.54} & $_{99}$\textbf{.61} & $_{99}$\textbf{.64} & $_{99}$\textbf{.70} & $_{99}$\textbf{.71} \\
\bottomrule
\end{tabular}
%\vspace{-6mm}
\end{table}

We first evaluate our method pre-quantisation on various modalities including images using CelebA-HQ dataset \citep{karras2018progressive}, manifolds using ERA5 \citep{hersbach2019era5}, 3d NeRF scenes using the SRN cars \citep{sitzmann2019srns} and 3d voxels using the top 10 classes of ShapeNet \citep{shapenet2015}. Following prior work, we train SIREN with 15 layers of 512 units and use MetaSGD \citep{li2017meta} with 3 inner-loop steps as our Meta-Learning method and use the same task batch size for comparable conditions. For baselines, we compare our technique with latent modulations using Functa \citep{dupont2022data} and sparse modulations using MSCN \citep{schwarz2022meta}. More details in the Appendix.

As illustrated in Table \ref{tab:modulation}, we demonstrates a marked improvement over previous approaches in almost all cases. Particularly noteworthy, \sname outperforms MSCN on ERA5 by more than 3.1dB when using a modulation size of 1024. This is particularly significant, as PSNR is based on a logarithmic scale. In addition, we note that the use of more complex latent $\rightarrow$ modulations/mask networks (as opposed to the linear projection of Functa) not only leads to better results, but also exhibits significantly faster learning progress (Figure \ref{fig:curves}).

A key hypothesis of our proposed conditioning technique is the idea of subnetwork selection. To provide empirical evidence and understand the behaviour of our conditioning method, we analyse the masks $\rmG_{\mathtt{low}}^{(l)}$ after obtaining  $\boldsymbol\phi^i$ for test set images. This is shown in Figure \ref{fig:modulation_analysis}. First, we note that product of gating masks and a shared, Meta-Learning matrix does indeed implement moderate sparsity levels (which we define as $|(\rmG_{\mathtt{low}} \odot \rmW)_{ij}| < 0.001$), despite avoiding the use of approximate inference (Figure \ref{fig:sparsity}). Remarkably, we observe sparsity levels varying significantly per layer, suggesting \sname learns \textit{where to learn}. This is particular significant as it is well known that only a fraction of layers typically need to adapted in Meta-Learning \citep{zintgraf2019fast}. Indeed, this was a key insight of MSCN \citep{schwarz2022meta} which we share despite our use of a much simpler sparsity/gating method.

Moreover, further examining our soft gating mechanism, we provide a t-SNE visualization \citep{maaten2008visualizing} of the adapted masks on CelebA-HQ (Figure \ref{fig:mask_celeba}). Resulting patterns intriguingly show clear clustering according to  image characteristics such as background color, indicating the ability to condition the shared INR based on image statistics.

\begin{figure*}[t]
\centering
\includegraphics[width=0.9\linewidth]{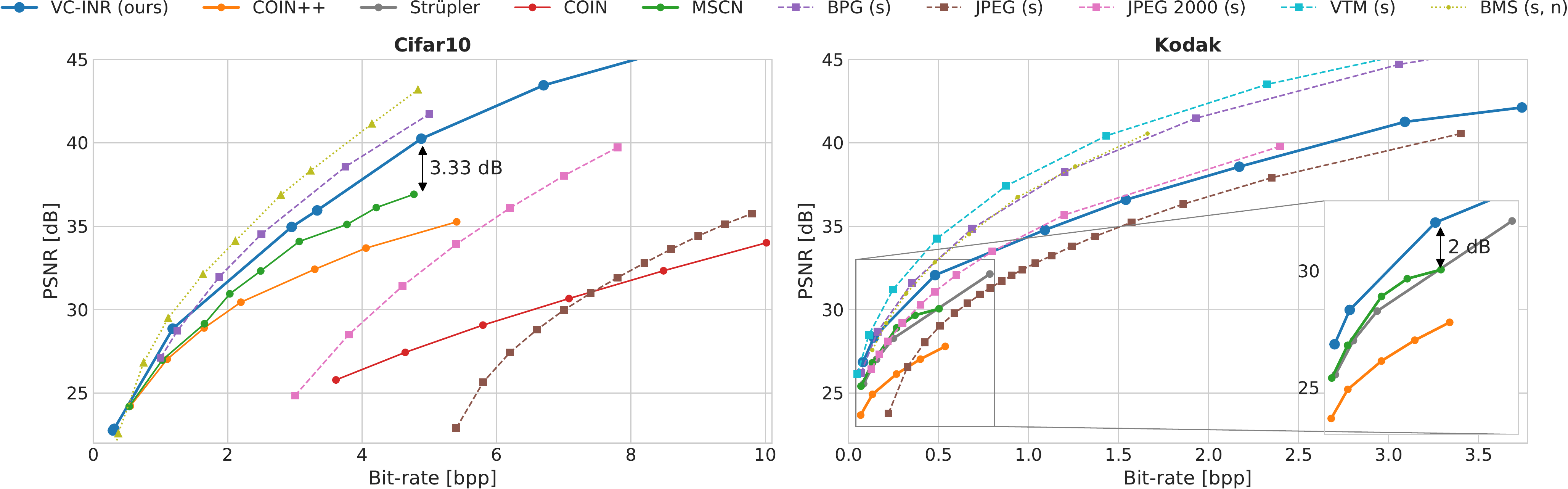}
\caption{Compression results on image datasets CIFAR-10 (left) \& Kodak (right). Modality-specific approaches are shown with a dashed line and marked \textit{(s)}. Conventional neural compression methods (also modality-agnostic) with a dotted line and marked \textit{(s, n)}. BMS is \citep{balle2017end}, Strüpler is \citep{strumpler2022implicit} and VTM \cite{bross2021overview}.}
\label{fig:images}
\end{figure*}

\begin{figure*}[t]
    \centering
    \begin{subfigure}{0.49\linewidth}
        \centering
        \includegraphics[width=\textwidth]{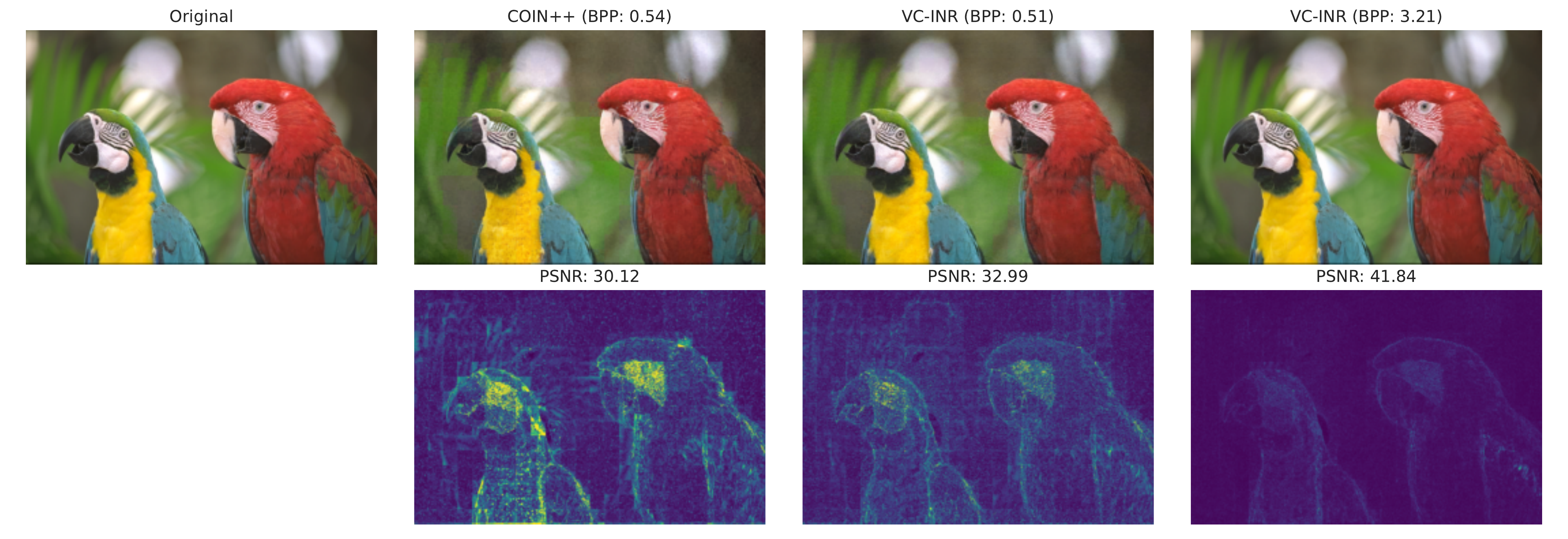}
        \caption{Compared with COIN++ \citep{dupont2022coinpp}.}
    \end{subfigure}
    \begin{subfigure}{0.49\linewidth}
        \centering
        \includegraphics[width=\textwidth]{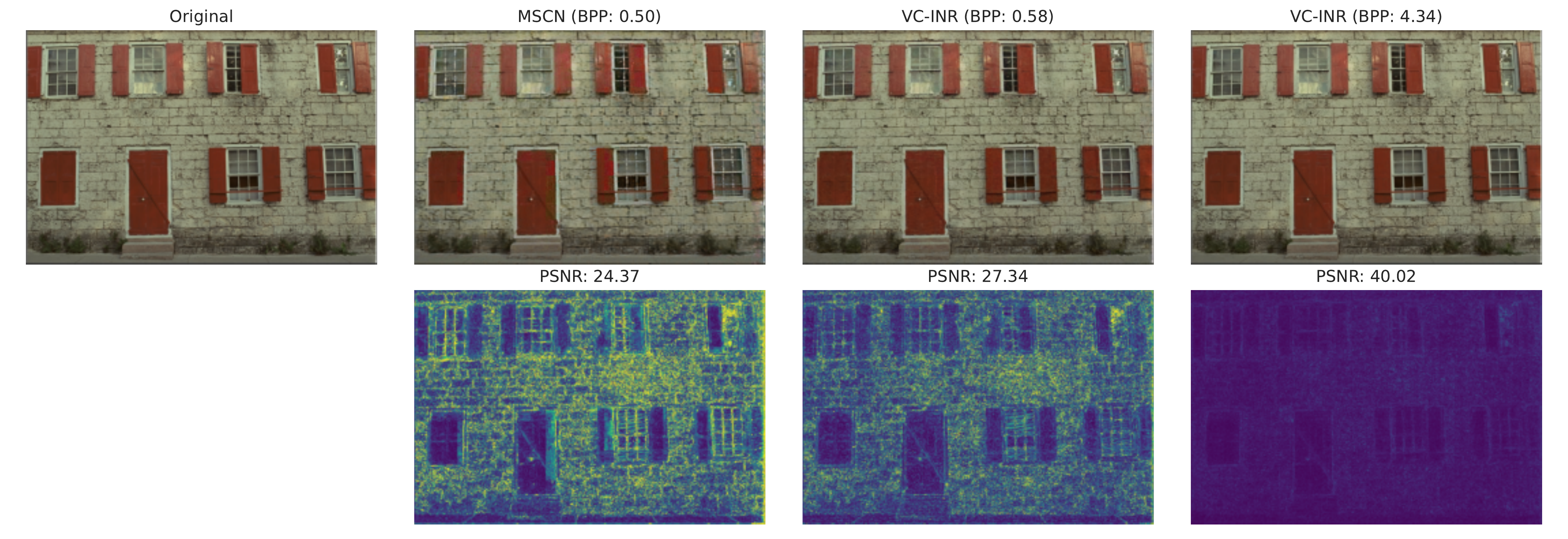}
        \caption{Compared with MSCN \citep{schwarz2022meta}.}
    \end{subfigure}
    %\begin{subfigure}{0.49\linewidth}
    %    \centering
    %    \includegraphics[width=\textwidth]{figures/kodak/vcinr_v_mscn_4.pdf}
    %    \caption{"}
    %\end{subfigure}
    \caption{Qualitative results from the Kodak dataset. Shown are VC-INR models in comparison with other INR-based techniques at similar bit-rates (3rd column) as well as a high-quality model (last column).}
    \label{fig:kodak_qualitative}
\end{figure*}

\begin{figure}[h]
    \centering
     \includegraphics[width=0.45\textwidth]{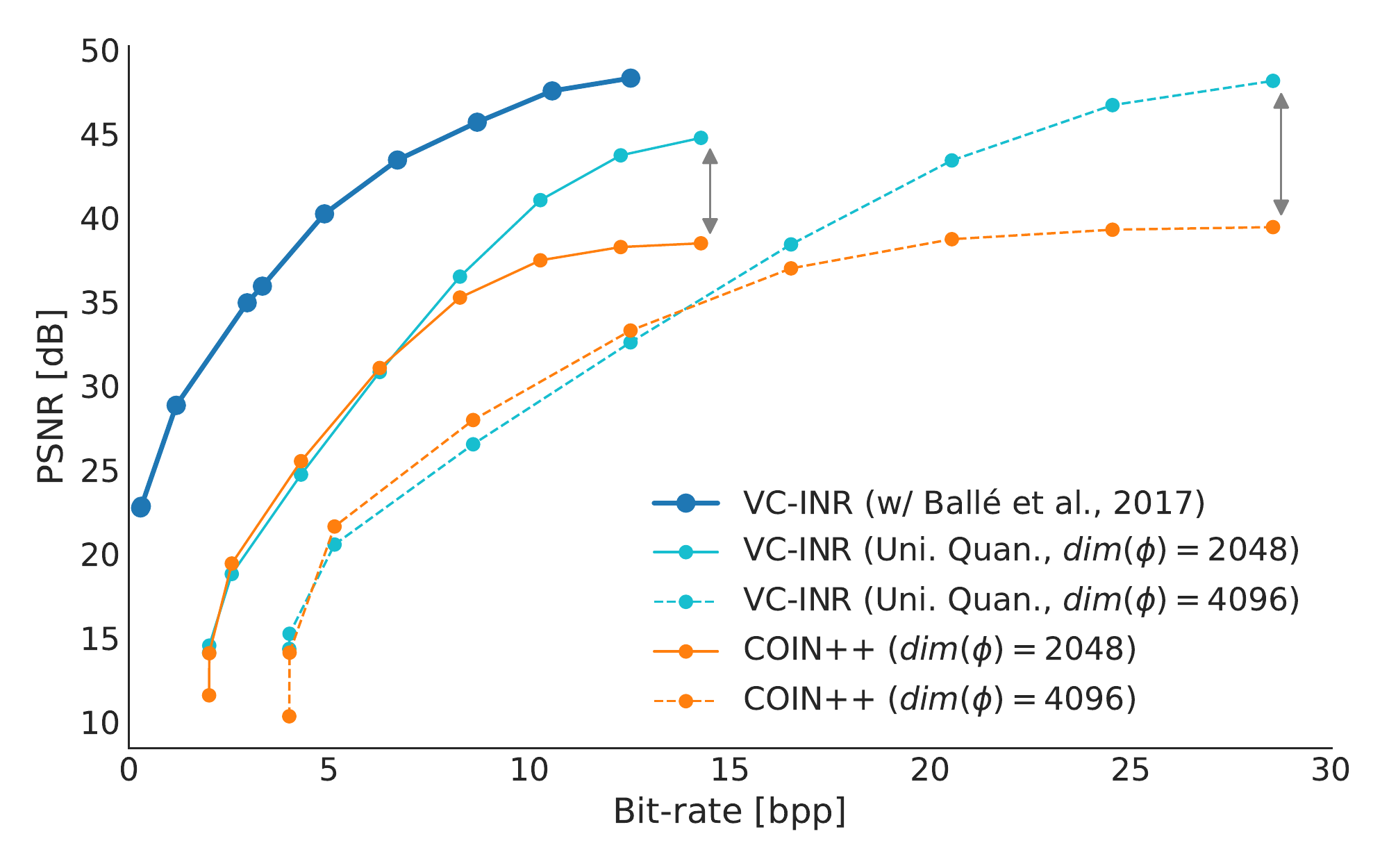}
    \caption{Learned vs uniform quantisation for VC-INR \& COIN++ on CIFAR-10.}
    \label{fig:uniform_v_balle}
\end{figure}

\subsection{Data compression across modalities}
\label{exp:compression}

We now evaluate \sname for data compression, the primary focus of our work. To demonstrate the versatility of \sname, we examine a range frequently encountered modalities. We measure reconstruction performance measured in terms of PSNR under different levels of compressed data sizes measured in kilobits per second (kbps) for audio and bits-per-pixel (bpp)\footnote{bpp$=\frac{\text{bits per parameters}\times\text{number of parameters}}{\text{number of pixels}}$}. Baselines are codecs such as JPEG \citep{wallace1992jpeg}, JPEG 2000 \citep{skodras2001jpeg}, BPG \citep{bellard2014bpg}, MP3 \citep{mp3codec}, AVC \citep{wiegand2003overview}, and HEVC \citep{bross2021overview}. We also compare against the modality-specific neural compression scheme BMS \citep{balle2018variational}, VTM \cite{bross2021overview} and other INR techniques such as COIN \citep{dupont2021coin}, COIN++ \citep{dupont2022coinpp}, MSCN \citep{schwarz2022meta} and the method in \citep{strumpler2022implicit}.

\textbf{Uniform vs Variational Compression}

While the previous section provides empirical justification for the architectural changes of VC-INR, we now additionally show the effectiveness of the proposed quantisation method. Figure \ref{fig:uniform_v_balle} contrasts rate-distortion curves obtained using Uniform Quantisation with the transform coding setup introduced earlier. Results are obtained by compressing the latent vectors obtained from two pre-trained models to varying bit-rates by varying the number of bits for uniform quantisation or the rate-distortion trade-off parameter $\lambda$ for the full VC-INR model. We note that the use of \citet{balle2017end}'s transform coding drastically shifts the rate-distortion curves towards lower bit-rates while maintaining a better reconstruction ratio. Furthermore, we can also see that the improved pre-training effectively increases the ceiling reconstruction performance when comparing uniform quantisation for VC-INR (light blue) with our COIN++ implementation (orange).

\textbf{Images} We show compression performance on the image domain using the CIFAR-10 \citep{krizhevsky2009learning} and Kodak (meta-trained on Div2k \citep{agustsson2017ntire}) datasets. In order to handle the large images found in the Kodak dataset, we divide the images into smaller patches as previously established in prior work.

Figure \ref{fig:images} shows that \sname significantly and consistently outperforms prior INR-based data compression methods (COIN, COIN++, MSCN, Str{\"u}mpler) and even certain image codecs (JPEG/JPEG 2000) on the CIFAR-10 dataset. In addition, \sname reconstruction continue to improve with higher bitrates, which we demonstrate by almost pixel-perfect reconstruction. This implies that learned entropy coding is a key factor in achieving strong results. Furthermore, \sname shows comparable performance to the strongest modality-specific methods at low bitrates, despite not taking advantage of inductive biases. While not fully matching state-of-the-art (SOTA) results on images compared to all compression techniques, we significantly reduce the gap. Note that we provide further results on Kodak using Multiscale structural similarity index measure (MS-SSIM) in the Appendix.

\begin{figure}[h]
\centering
    \includegraphics[width=\linewidth]{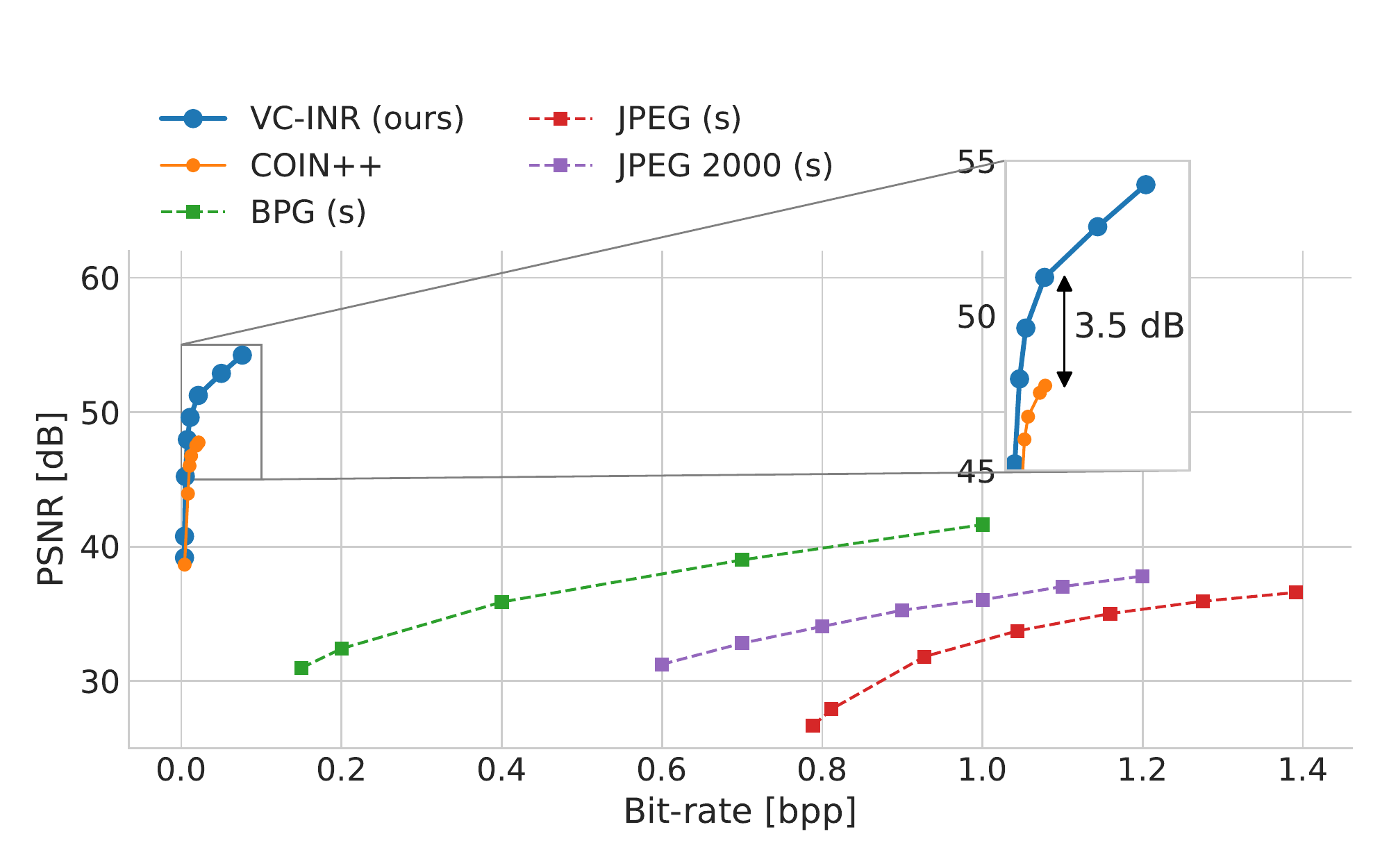}
    \caption{Compression results on ERA5 (climate data/manifolds).}
    % Methods marked (s) are modality-specific.}
    \label{fig:era5}
\end{figure}

\textbf{Manifolds} This evaluates \sname on global temperature measurements from the ERA5 ($16\times$) dataset. The dataset consists of temperature measurements (features) at equally spaced latitudes and longitudes (coordinates) on Earth from 1979 to 2020, represented by spherical coordinates. Since no codec or neural compression method has been developed specifically for this modality, we compare \sname to COIN++ and image codecs (applied by unrolling the manifold on a rectangular grid) as baselines. As shown in Figure \ref{fig:era5}, \sname, we achieve an improvement of approximately 3.5dB at the same bitrate compared to the SOTA. This highlights the large potential impact modality-agnostic techniques might have for specialised data types.

\begin{figure}[h]
\centering
\begin{subfigure}{0.5\textwidth}
\centering
    \includegraphics[width=0.85\linewidth]{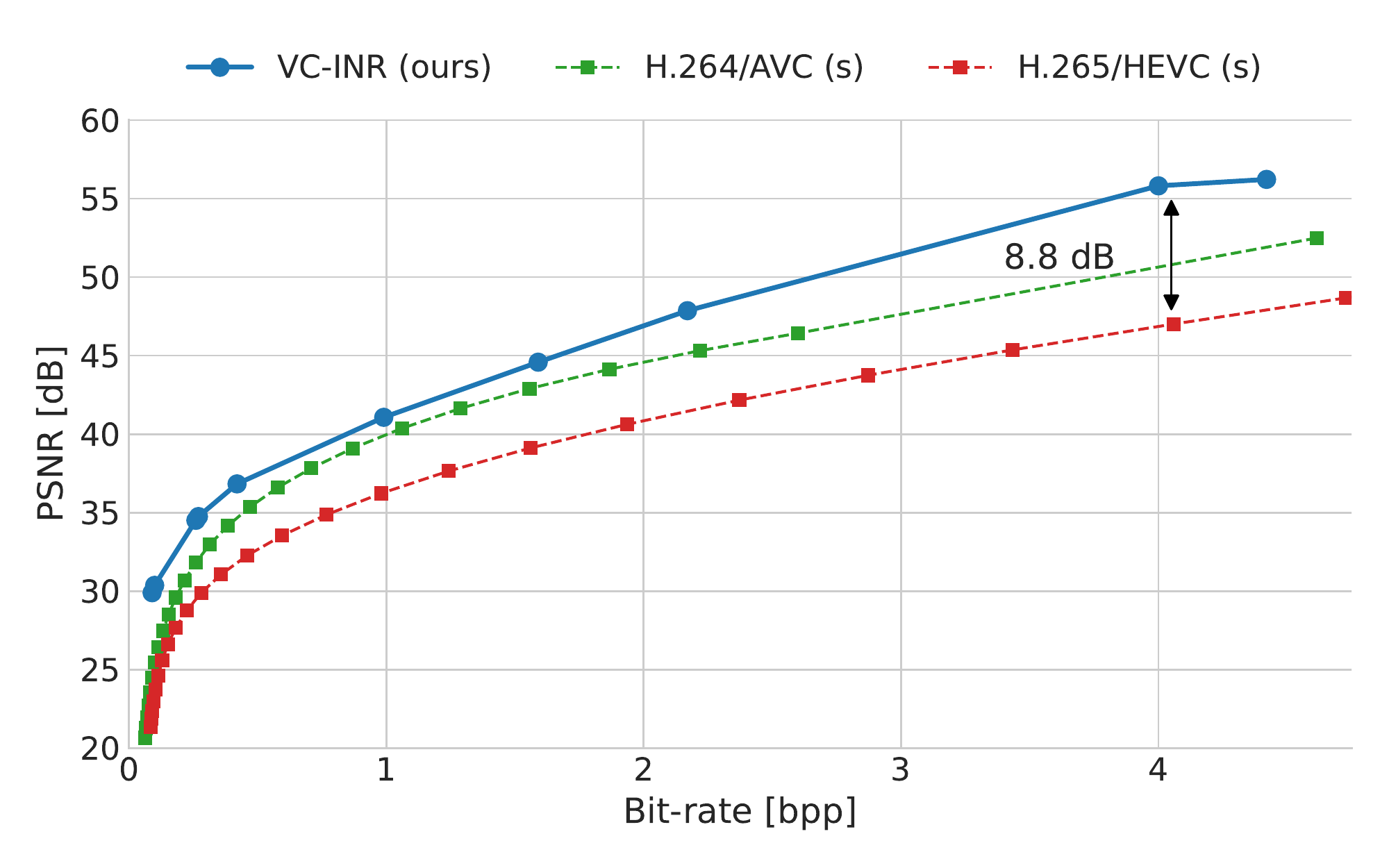}
    \caption{UCF-101}
    \label{fig:ucf}
\end{subfigure}%
\newline
\begin{subfigure}{0.5\textwidth}
\centering
    \includegraphics[width=0.85\linewidth]{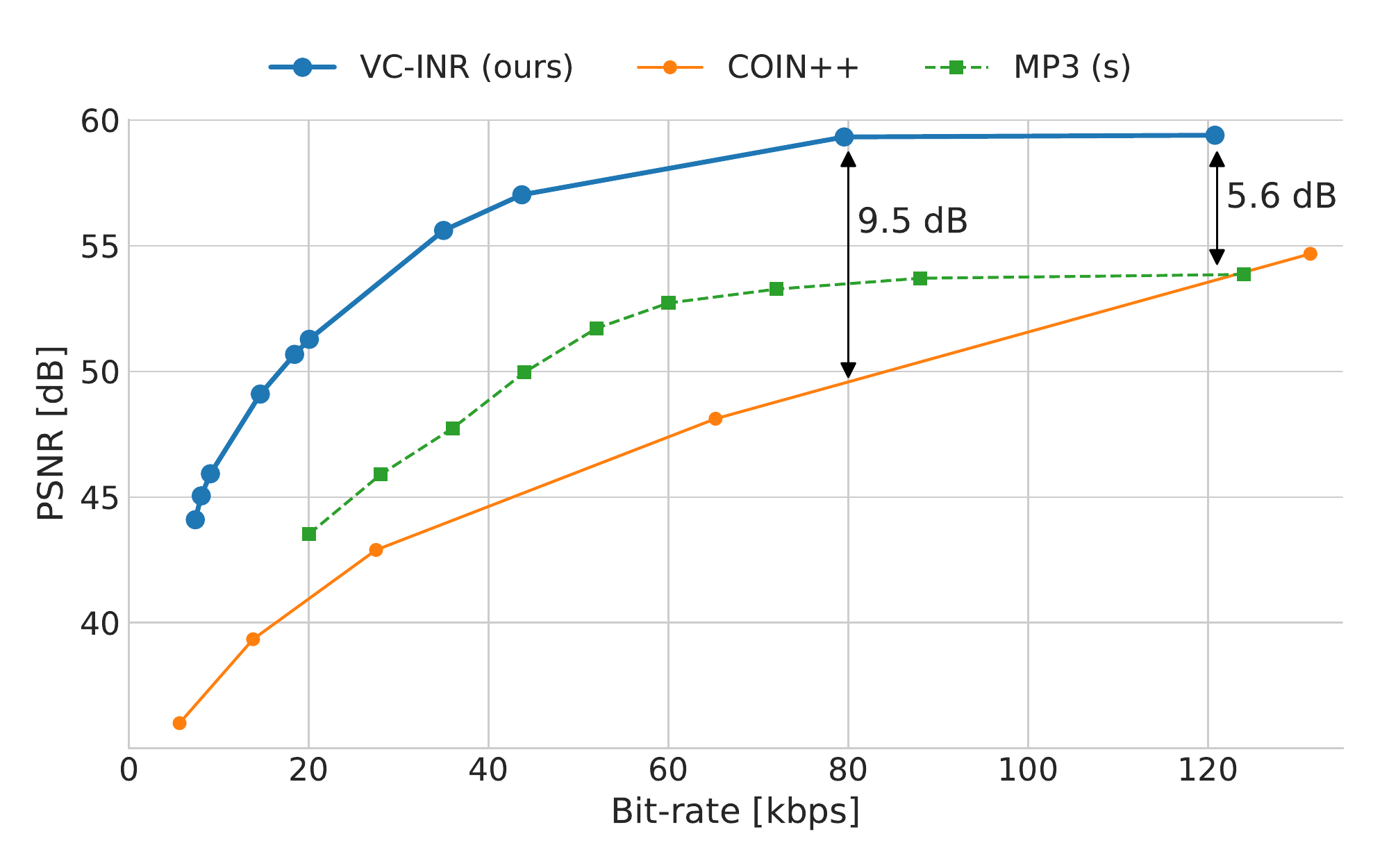}
    \caption{LibriSpeech}
    \label{fig:libri}
\end{subfigure}
\caption{Compression results on (a) videos and (b) audio.}
\label{fig:climate_video}
\end{figure}

\textbf{Audio} Evaluating \sname in the audio domain, we utilise the LibriSpeech dataset \citep{panayotov2015librispeech}, a large speech dataset recorded at a 16kHz sampling rate. We consider the MP3 codec as well as COIN++ as baseline methods. As in COIN++ we use patching to keep the comparison fair. Our results, shown in Figure \ref{fig:libri} demonstrate impressive results, showing that \sname significantly outperforms both COIN++ as well as the widely used and popular MP3 codec.

\begin{figure}[h]
\centering
    \includegraphics[width=0.8\linewidth]{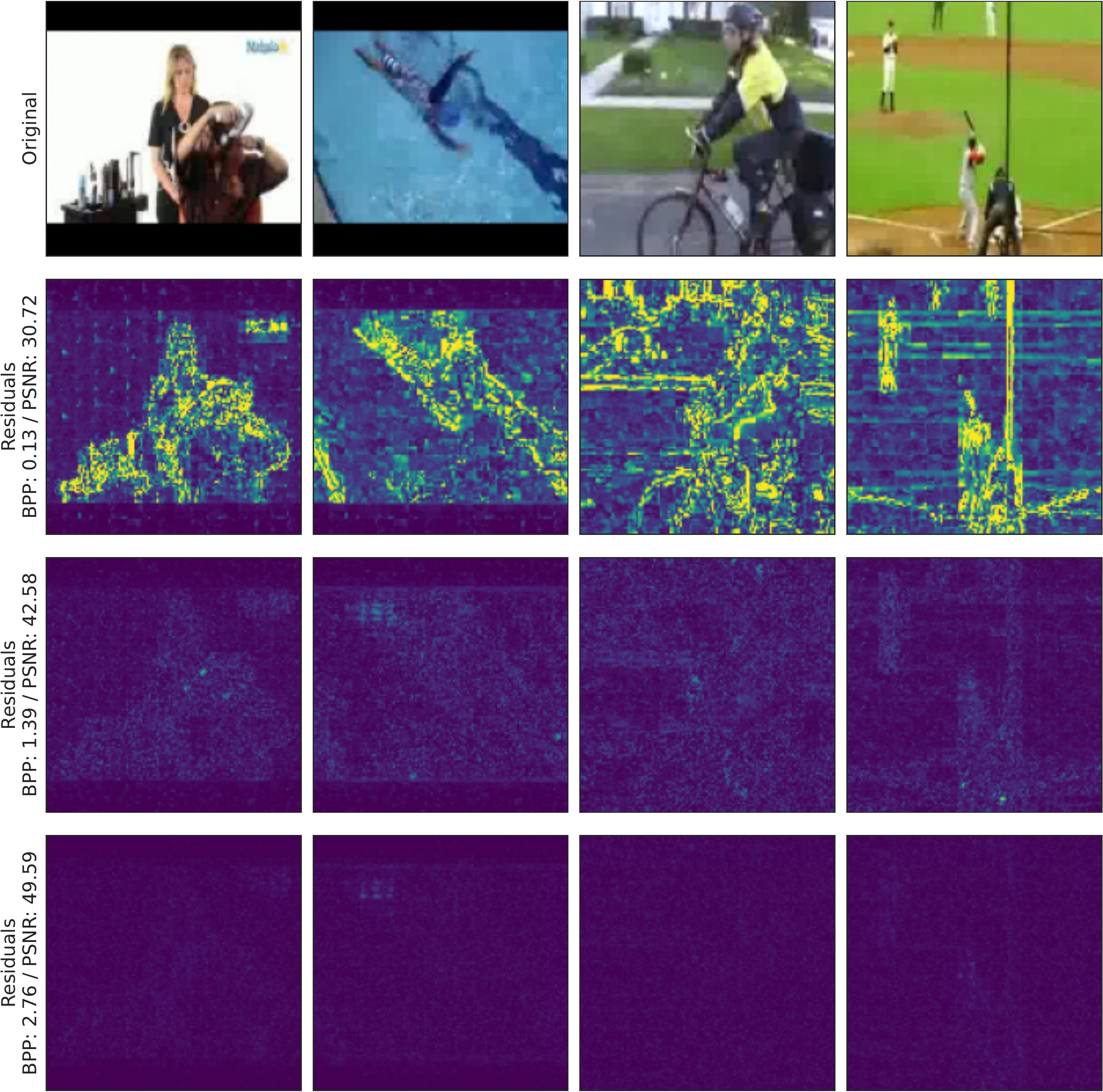}
    \caption{Results on videos showing residuals of VC-INR at various quality levels. Videos available: \href{https://drive.google.com/file/d/1Z6RgEZjeBALsLE0UxZ0Xz6t0VsNKAvrV/view?usp=sharing}{0.13 bpp}, \href{https://drive.google.com/file/d/1-EWmFK5qw4FWLb4KpERSzFv56vr7tCbR/view?usp=sharing}{1.39 bpp}, \href{https://drive.google.com/file/d/1mnAZWTG6wE_p1-9j8PRN9I7uIrOodCWn/view?usp=sharing}{2.76 bpp}.}
    % Methods marked (s) are modality-specific.}
    \label{fig:ucf_qualitative}
\end{figure}

\textbf{Videos} Turning to the video domain, we compress clips from the UCF-101 action recognition dataset \citep{soomro2012ucf101}, once again using patching. Here, we compare \sname to video codecs AVC and HEVC. Impressively, \sname outperforms both, raising hopes for the potential of INR-based compression to one day replace hand-designed codecs for video. Qualitative results are available in Figure \ref{fig:ucf_qualitative}, showing the prediction errors of \sname models at varying bit-rates, for all of which we achieve SOTA better results.

% ==========================================================
\section{Conclusion}

We introduce VC-INR, a modality-agnostic neural compression technique showing strong and consistent improvements over previous INR-based methods. This was achieved by developing algorithmic improvements across the two axes of representational power and advanced quantisation while maintaining modality-agnosticism. Our technique bridges the gap between recent approaches to compact INR representations based on latent codes and sparsity and shows how previously modality-specific algorithms can be elevated to the modality-agnostic setting. Our evaluation shows strong improvement on previous work with INRs \citep[e.g.][]{dupont2022coinpp, schwarz2022meta} while outperforming certain established algorithms (e.g. JPEG on images, MP3 on audio and AVC/HEVC on videos) while reducing the gap to others (e.g. BPG or BMS on images). We believe that the conceptual advantage of a single algorithm applicable to all modulations will continue to show rapid improvements as innovations are developed. Future work may focus on including further developments such as advanced priors \citep[e.g.][]{balle2018variational, cheng2020learned, ladune2022cool}. In addition, improved patching strategies resulting from e.g. memory-efficient Meta-Learning algorithms or path allocation based on signal variation might be fruitful.
% ==========================================================

% \section*{Acknowledgements}
% \input{section/acknowledgements}

\bibliography{main}
\bibliographystyle{icml2023}

%%%%%%%%%%%%%%%%%%%%%%%%%%%%%%%%%
% APPENDIX
%%%%%%%%%%%%%%%%%%%%%%%%%%%%%%%%%
\newpage
\appendix
\onecolumn
\section{Dataset description}
\label{sec:appendix-dataset}

\textbf{CelebA-HQ} is a high-quality version of the CelebA dataset, which includes images of celebrities along with corresponding attributes \citep{karras2018progressive}. By following \citep{dupont2022data}, we divide the dataset into 27,000 training examples and 3,000 test examples, and pre-processed the pixel coordinates into $[0, 1]^2$ and feature values ranging from 0 to 1.

\textbf{ShapeNet} is a dataset of 3D shapes of 10 different object categories \citep{shapenet2015}. We follow the pre-processing by \citep{dupont2022data}, and downscale the resolution of 128$^3$ to 64$^3$ by using \texttt{scipy.ndimage.zoom} function with threshold 0.05. To augment the datasets, the authors applied a 50-fold expansion by independently scaling the shapes in the x, y, and z axes using a randomly sampled scale within the range of 0.75 to 1.25. The resulting dataset includes 1,516,750 training examples and 168,850 test examples with voxel coordinates into $[0, 1]^3$ and occupancies in binary \{0, 1\}.

\textbf{ERA5} is a dataset consists of temperature observations from 1979 to 2020 on a global grid of equally spaced latitudes and longitudes \citep{hersbach2019era5}. By following \citep{dupont2022data}, we downsample the grid resolution 721 $\times$ 1044 to 181 $\times$ 360. Each time step is treated as a separate data point, and the dataset is split into a training set of 9676 data points and a test set of 2420 data points. As for the input, the given latitudes $c_{\mathtt{lat}}$ and longitudes $c_{\mathtt{long}}$ are transformed into 3D Cartesian coordinates $\rvc=$ ($\cos c_{\mathtt{lat}}$ $\cos c_{\mathtt{long}}$, $\cos c_{\mathtt{lat}}$ $\sin c_{\mathtt{long}}$, $\sin c_{\mathtt{lat}}$) where latitudes $c_{\mathtt{lat}}$ are equally spaced in $[-\frac{\pi}{2} , \frac{\pi}{2}]$ and longitudes $c_{\mathtt{long}}$ are equally spaced in $[0 , \frac{2\pi(n-1)}{n}]$ where $n$ the number of distinct values of longitude (360).

\textbf{SRN Cars} is a dataset of car scenes, with 2458 examples in the training set and 703 examples in the test set \citep{sitzmann2019srns}. Each example consists of 50 random views centered on the car in the training set, and 251 views in the test set. The pre-processing of the data was conducted according to the guidelines provided by \citep{dupont2022coinpp}.

\textbf{CIFAR-10} is a dataset of 50,000 train and 10,000 test images with a resolution of 32 x 32, comprising 10 different object categories \citep{krizhevsky2009learning}. We use the same pre-processing as in CelebA-HQ dataset.

\textbf{Kodak} is a dataset of 24 uncompressed PNG images with a resolution of 768 $\times$ 512, provided by the Kodak corporation. By following \citep{schwarz2022meta}, we meta-learn on the high-quality versions of the Div2K dataset \citep{agustsson2017ntire}, which consists of 900 images (by combining train and validation set). For Meta-Learning, we also train the model on randomly cropped 32 $\times$ 32 patches and for evaluation, we split the image into non-overlapping patches where each modulations are adapted on each patches. Here, we also use the same pre-processing as in CelebA-HQ dataset.

\textbf{LibriSpeech} is a collection of read English speech recordings at a 16kHz sampling rate \citep{panayotov2015librispeech}. By following \citet{dupont2022coinpp}, we use the train-clean-100 split, which consists of 28,539 examples, and the test-clean split, which consists 2,620 examples. For the experiments, we use the first 3 seconds of each example (which is 48,000 audio samples) for both training and evaluation. For the pre-processing, we scale the coordinates into $[-5, 5]$.

\textbf{UCF-101} is a video action dataset comprising 13,320 videos with a resolution of 320 $\times$  240, organised into 101 classes \citep{soomro2012ucf101}. In order to standardise the input for the model, we center-crop each video clip to 240 $\times$ 240 $\times$ 24 and then resized to 128 $\times$ 128 $\times$ 24.
% ==========================================================
\section{Numerical results}
\label{sec:appendix-numerical results}

For the sake of reputability, we now state the numerical compression values used to plot the results in Section \ref{sec:exp}. Note that baseline results have been taken from the \href{https://github.com/EmilienDupont/coinpp/tree/main/results}{code repository for COIN++} \citep{dupont2022coinpp}:

\begin{small}
\begin{verbatim}
# Cifar-10
vcinr_bpp = [0.29, 0.31, 1.18, 1.18, 2.95, 3.33, 4.88, 6.70, 8.69, 10.56, 12.52]
vcinr_psnr = [22.76, 22.86, 28.86, 28.86, 34.96, 35.95, 40.25, 43.45,  45.70, 47.56, 48.32]

# Kodak
vcinr_bpp = [0.08, 0.14, 0.48, 1.09, 1.54, 2.17, 3.09, 3.74, 5.56]
vcinr_psnr = [26.86, 28.33, 32.07, 34.78, 36.59, 38.57, 41.26, 42.12, 42.24]

# ERA-5
vcinr_bpp = [0.004, 0.004, 0.005, 0.00758, 0.011, 0.02119, 0.05, 0.07616]
vcinr_psnr = [39.172, 40.766, 45.219, 47.965, 49.612, 51.25, 52.89, 54.25]

# Librispeech
vcinr_bpp = [7.38, 8.04, 9.06, 14.61, 18.42, 20.06, 34.99, 43.69, 79.54, 120.77]
vcinr_psnr = [44.10, 45.05, 45.93, 49.10, 50.68, 51.28, 55.61, 57.03, 59.33, 59.40]

#  UCF-101
vcinr_bpp = [0.09, 0.10, 0.26, 0.27, 0.42, 0.99, 1.59, 2.17, 4.00, 4.42]
vcinr_psnr = [29.90, 30.37, 34.51, 34.75, 36.83, 41.07, 44.58, 47.86, 55.81, 56.22]
\end{verbatim}
\end{small}
% ==========================================================
\section{Meta-Learning implicit neural representations with latent modulations}
\label{sec:appendix-meta}

In order to efficiently and effectively encode a given signal into a compact latent representation, we utilise a Gradient-based Meta-Learning approach, such as model-agnostic meta-learning (MAML) \citep{finn2017model}. In our case, MAML aims to find a good initialisation $\boldsymbol\phi_{0}$ and shared INR parameter $\boldsymbol\theta$, allowing for the encoding of a given signal $\rvx$ into the modulation $\boldsymbol\phi$ within a few gradient steps from $\boldsymbol\phi_{0}$. 
Writing $\mathcal{L}_{\mathtt{MSE}}(\boldsymbol\theta,\boldsymbol\phi, \rvx)$ as a shorthand for the INR fitting loss (Equation (\ref{eq:mse-inr})), a single gradient step adaptation of MAML is computed as:
\begin{equation}
    \boldsymbol\phi = \boldsymbol\phi_{0} - \alpha \nabla_{\boldsymbol\phi_0} \mathcal{L}_{\mathtt{MSE}}(\boldsymbol\theta,\boldsymbol\phi_0, \rvx),
    \label{eq:maml_in}
\end{equation}
where $\alpha$ is the step size used in the inner loop. Note that one can easily iterate the adaptation for multiple steps. The key idea of MAML is to backpropagate through this optimisation process, directly learning an initialisation $\boldsymbol\phi_0$ (along with additional shared parameters $\boldsymbol\theta$) such that $\boldsymbol\phi$ can parameterise a good reconstruction of the signal after adaptation. This is typically computed over the training signal distribution $p(\rvx)$:
\begin{equation}
    \min_{\boldsymbol\theta,\boldsymbol\phi_0} \mathbb{E}_{\rvx \sim p(\rvx)} \Big[ \mathcal{L}_{\mathtt{MSE}}(\boldsymbol\theta,\boldsymbol\phi, \rvx)\Big] = \min_{\boldsymbol\theta,\boldsymbol\phi_0} \mathbb{E}_{\rvx \sim p(\rvx)} \Big[ \mathcal{L}_{\mathtt{MSE}}\big(\boldsymbol\theta,\boldsymbol\phi_{0} - \alpha \nabla_{\boldsymbol\phi_0} \mathcal{L}_{\mathtt{MSE}}(\boldsymbol\theta,\boldsymbol\phi_0, \rvx), \rvx\big) \Big].
    \label{eq:maml_out}
\end{equation}
Here, we refer each optimisation of MAML, Equation (\ref{eq:maml_in}) as ``inner-loop'', and (\ref{eq:maml_out}) as ``outer-loop'', respectively. In practise, we also meta-learn the step size $\alpha$ as another parameter updated in the outer loop, an approach known as MetaSGD \citep{li2017meta}. This can be interpreted as a pre-conditioning of the gradient.

% ==========================================================
\begin{minipage}[t]{0.475\textwidth}
\vspace{0pt}  
\begin{algorithm}[H]
 %\SetKwInOut{Input}{Input}\SetKwInOut{Output}{Output}
\KwData{Dataset $\{\rvx^i, \rvy^i\}_{i=1}^N$}
 Initialise shared network $\boldsymbol\theta$ and latent modulation initialisation $\boldsymbol\phi_0$.\\

\While{not converged}{
     Sample batch of data $\mathcal{B} = \{\rvx^j, \rvy^j\}_{j=1}^B$\\
      \tcp{Adaptation loop ($\mathcal{O}$ in Figure \ref{fig:operational_diagram_c})}
      \For{$j\leftarrow 1$ \KwTo $B$}
      {
        \tcp{For 1 adaptation step}
         $\boldsymbol\phi^j  \leftarrow \boldsymbol\phi_0 - \alpha\nabla_{\boldsymbol\phi_0}\mathcal{L}_{\texttt{MSE}}(f(\rvx^j, \boldsymbol\theta, \textcolor{red}{\boldsymbol\phi_0}), \rvy^j)$\\
      }
   \tcp{Update using adapted latent modulation}
   $\boldsymbol\phi_0 \leftarrow \boldsymbol\phi_0 - \beta \mathbb{E}[\nabla_{\boldsymbol\phi_0}\mathcal{L}_{\texttt{MSE}}(f(\rvx^j, \boldsymbol\theta, \textcolor{red}{\boldsymbol\phi^j}), \rvy^j)]$\\
   \tcp{Remaining INR parameters}
   $\boldsymbol\theta  \leftarrow \boldsymbol\theta - \beta \mathbb{E}[\nabla_{\boldsymbol\theta}\mathcal{L}_{\texttt{MSE}}(f(\rvx^j, \boldsymbol\theta, \textcolor{red}{\boldsymbol\phi^j}), \rvy^j)]$\\
}
 
\KwResult{Dataset of latent modulations $\{\boldsymbol\phi^i\}_{i=1}^N$, $\boldsymbol\theta$}
\caption{INR Meta-training stage}
\label{algo:pretrain}
\end{algorithm}
\end{minipage}%
\hfill
\begin{minipage}[t]{0.475\textwidth}
  \vspace{0pt}
\vspace{0pt}  
\begin{algorithm}[H]
 %\SetKwInOut{Input}{Input}\SetKwInOut{Output}{Output}
\KwData{Dataset of latent modulations $\{\boldsymbol\phi^i\}_{i=1}^N$, $\boldsymbol\theta$, $\lambda$}

\While{not converged}{
    Initialise parameters $\boldsymbol\pi_a, \boldsymbol\pi_s$.\\
     Sample batch of data $\mathcal{B} = \{\boldsymbol\phi^j, \rvx^j, \rvy^j\}_{j=1}^B$\\
      \For{$j\leftarrow 1$ \KwTo $B$}
      {
         %\tcp{Analysis transform}
         $\rvz \leftarrow g_a(\boldsymbol\phi^j; \boldsymbol\pi_a)$\\
         \tcp{\textcolor{blue}{Rounding at inference to obtain $\hat{\rvz}^j$}}
          $\widetilde{\rvz}^j = \rvz^j + \boldsymbol\epsilon; \boldsymbol\epsilon \sim \mathcal{U}(-\frac{1}{2},\frac{1}{2})$\\
         \tcp{Compute entropy model $p_{\hat{\rvz}}$ and rate}
         $\ell^j_{\mathtt{rate}} =-\log_2[p_{\hat{\rvz}}(\widetilde{\rvz})]$\\
         %\tcp{Synthesis transform}
         $\widetilde{\boldsymbol\phi}^j \leftarrow g_s(\widetilde{\rvz}^j; \boldsymbol\pi_s)$\\
         $\ell^j_{\mathtt{distortion}} =\mathcal{L}_{\texttt{MSE}}(f(\rvx^j, \boldsymbol\theta, \widetilde{\boldsymbol\phi}^j), \rvy^j)$
      }
   $\boldsymbol\pi_a \leftarrow \boldsymbol\pi_a - \beta \mathbb{E}[\nabla_{\boldsymbol\pi_a}(\ell^j_{\mathtt{rate}} + \lambda \ell^j_{\mathtt{distortion}})]$\\
   $\boldsymbol\pi_s \leftarrow \boldsymbol\pi_s - \beta \mathbb{E}[\nabla_{\boldsymbol\pi_s}(\ell^j_{\mathtt{rate}} + \lambda \ell^j_{\mathtt{distortion}})]$\\
}
 
%\KwResult{$\boldsymbol\pi}
\caption{Quantisation training stage}
\label{algo:compress}
\end{algorithm}
\end{minipage}

\section{VC-INR algorithmic details}
\label{sec:appendix-algo}

Algorithms \ref{algo:pretrain} and \ref{algo:compress} show details of the Meta-Learning (introduced in the previous section) and quantisation learning stages. The output of Algorithm \ref{algo:pretrain} directly feeds into the pipeline for quantisation. Hence, the two problems of optimal parameterisation and quantisation can be tackled independently, thus allowing for various combination for future work. 
\section{Additional experimental results}

\begin{figure*}[t]
\centering
%
%\begin{subfigure}{0.33\textwidth}
%    \centering
%    \includegraphics[width=\linewidth]{figures/cifar10_tsne_small.png} 
%    \caption{Mask clustering on Cifar10}
%    \label{fig:Cifar10 TSNE}
%\end{subfigure}
%
\begin{subfigure}{0.45\textwidth}
    \centering
    \includegraphics[width=\linewidth]{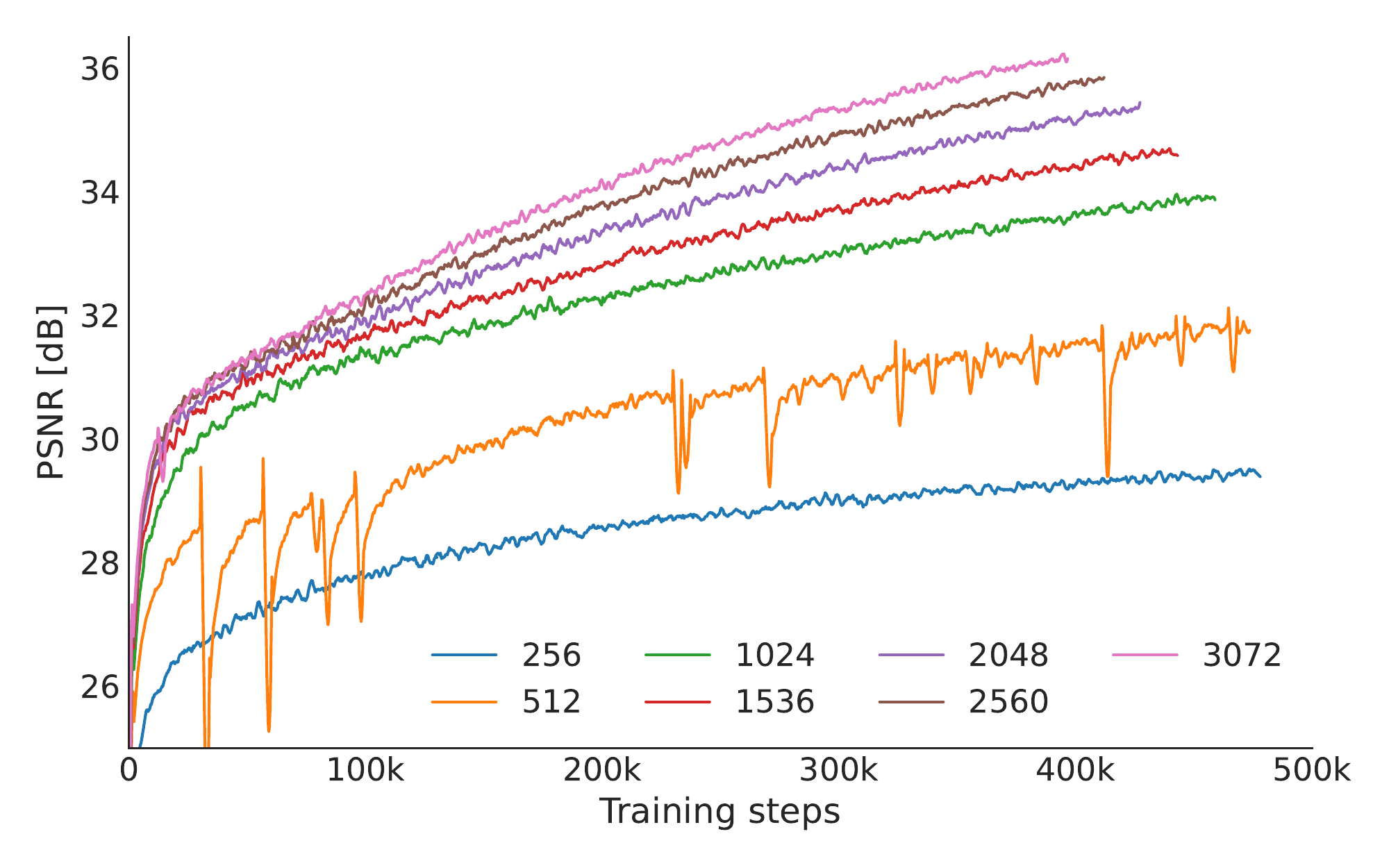} 
    \caption{Effect of the projection layer width size}
    \label{fig:non-linear-width}
\end{subfigure}
\qquad
\begin{subfigure}{0.45\textwidth}
    \centering
    \includegraphics[width=\linewidth]{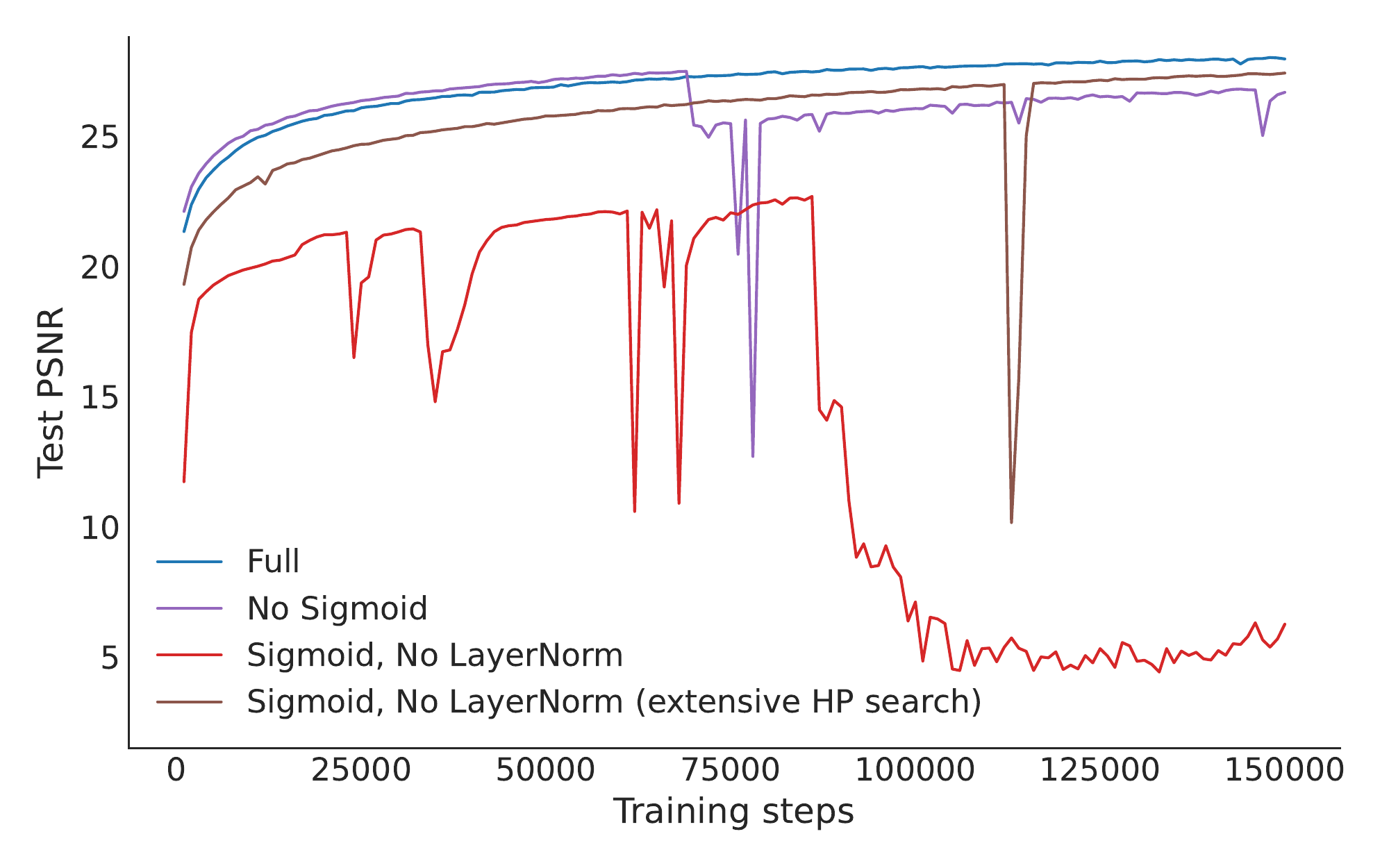} 
    \caption{Effect of LayerNorm}
    \label{fig:layernorm}
\end{subfigure}
\caption{Performance during the meta-training phase. (a) investigates the effect of the width of the \sname non-linear projection layer and (b) compares the effect of LayerNorm on \sname.}
\label{fig:ablation_study}
\end{figure*}

\subsection{Stabilising Meta-Learning of soft gating mask modulations with LayerNorm}

In Section \ref{sec:method-low-rank}, we demonstrate the importance of using LayerNorm \citep{ba2016layer} in the meta-learning of our new parameterisation. In Figure \ref{fig:layernorm} we demonstrate that Meta-Learning becomes highly unstable by default (an effect becoming more severe with larger $\dim(\boldsymbol\phi)$) and thus requires extensive hyperparameter search which may still suffer from occasional instability. Instead, we find that LayerNorm largely removes this phenomenon, leading to more stable training and better results. We hypothesise that such a divergence occurs when the norm of the inner loop gradient is large, indicating a sharp loss landscape. LayerNorm addresses this issue by smoothing the loss landscape, as has previously been shown \citep{santurkar2018does,xu2019understanding}. Furthermore, it effectively bounds the norm of $\phi$.

In addition, we show that our new parameterisation can effectively make use of increasing network capacity (while Functa shows decreasing performance for non-linear mappings from latent parameters to modulations). Figure \ref{fig:non-linear-width} shows this effect to be particularly effective for increasing network width, which we recommend for optimal performance during pre-training. 

\subsection{Results using MS-SSIM}

\begin{figure}[h]
\centering
\begin{subfigure}{0.45\textwidth}
\centering
    \includegraphics[width=\linewidth]{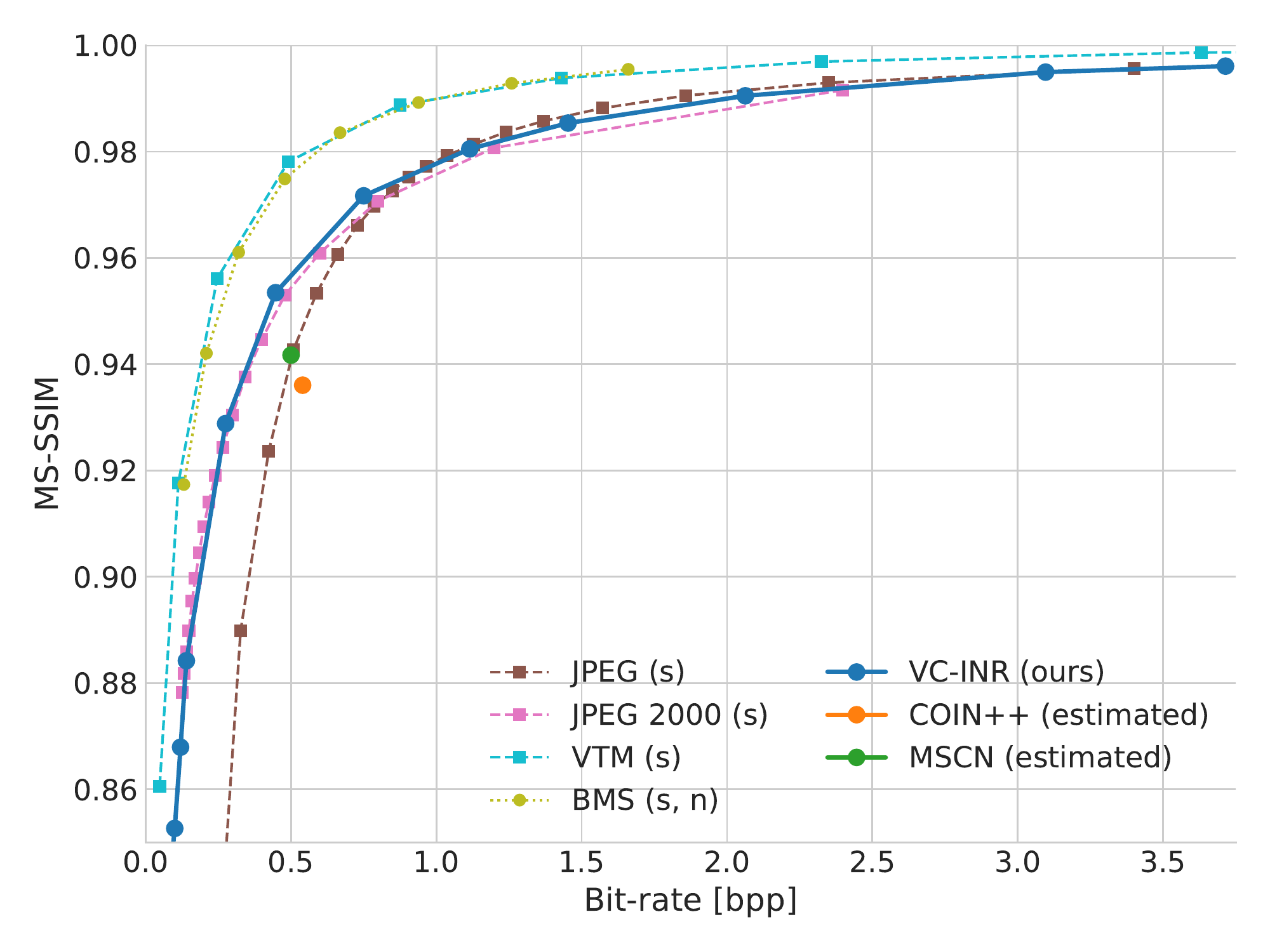}
\end{subfigure}%
\qquad
\begin{subfigure}{0.45\textwidth}
\centering
    \includegraphics[width=\linewidth]{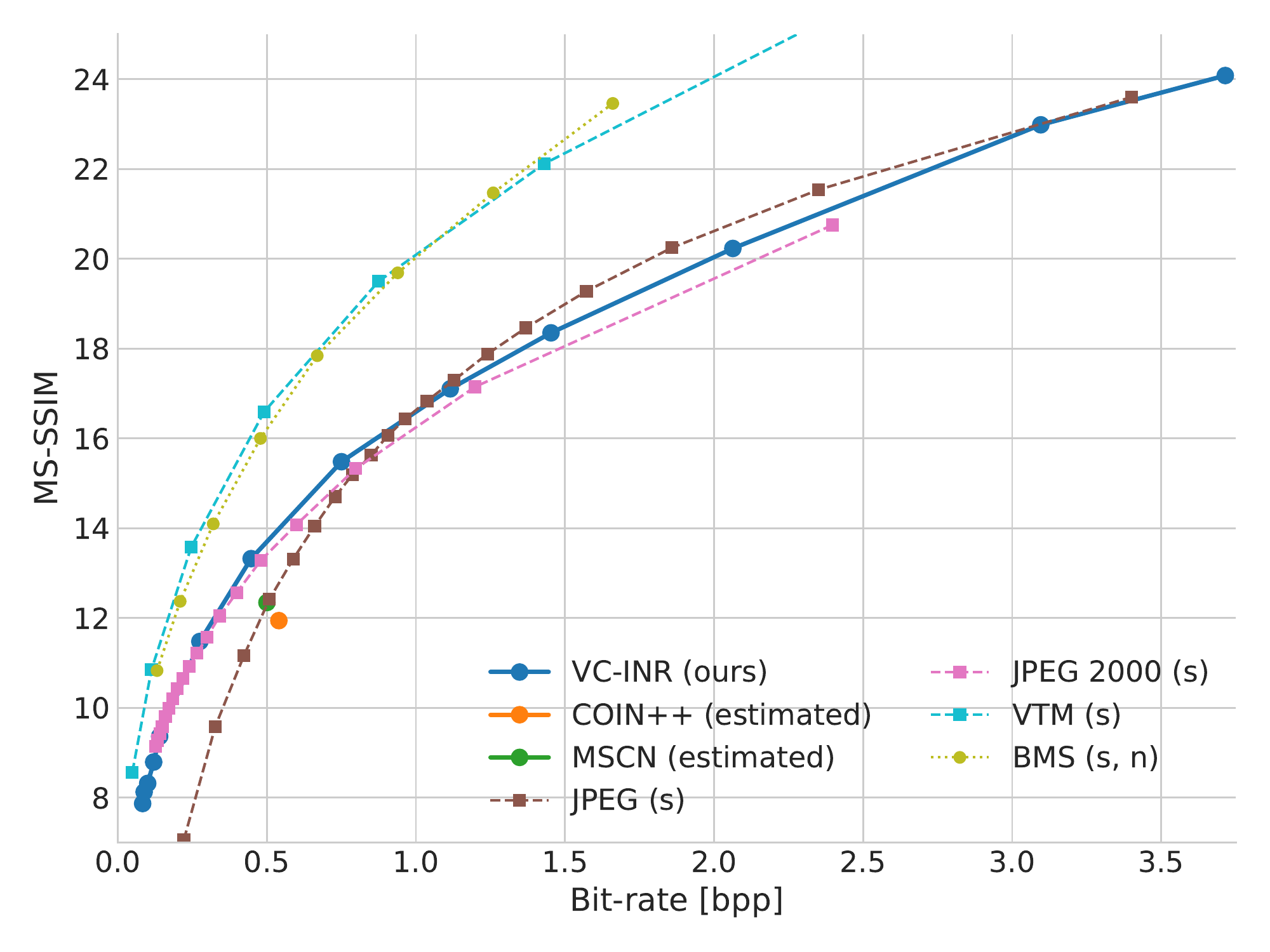}
\end{subfigure}
\caption{Compression results on image dataset Kodak measured using Multi-Scale Structural Similarity (MS-SSIM). (a) MS-SSIM scores, (b) Converted to decibel (i.e. $-10\log_{10}(1-\text{MS-SSIM})$).}
\label{fig:kodak_msssim}
\end{figure}

In addition to results measured using PSNR, we provide results on Kodak using Multiscale structural similarity index measure (SSIM) \citep{wang2003multiscale} results in Figure \ref{fig:kodak_msssim} due to its better correlation with perceptual similarity. We observe comparable the results in Figure \ref{fig:images} with VC-INR performing similarly to JPEG \& JPEG-2000.

\section{Qualitative Results}
\subsection{Cifar10}

\begin{figure}[h]
    \centering
    \begin{subfigure}{0.49\linewidth}
        \centering
        \includegraphics[width=\textwidth]{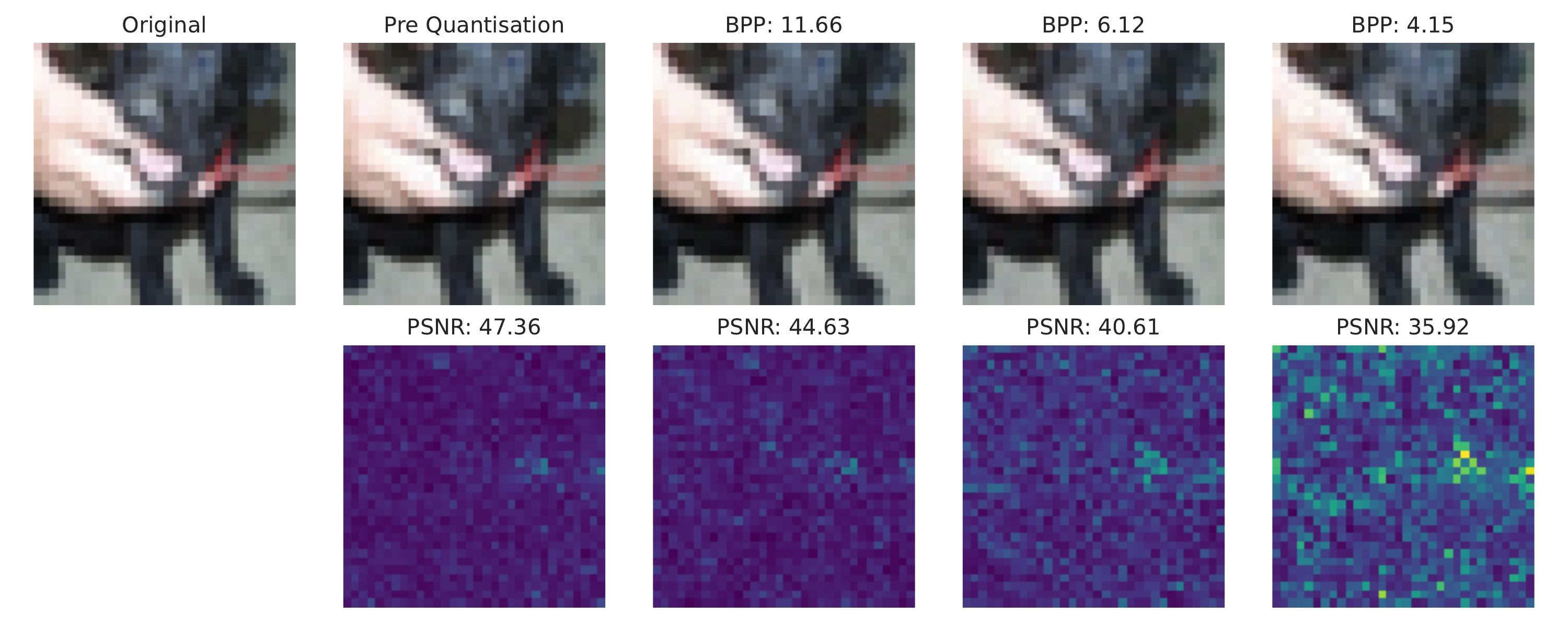}.
    \end{subfigure}
    \begin{subfigure}{0.49\linewidth}
        \centering
        \includegraphics[width=\textwidth]{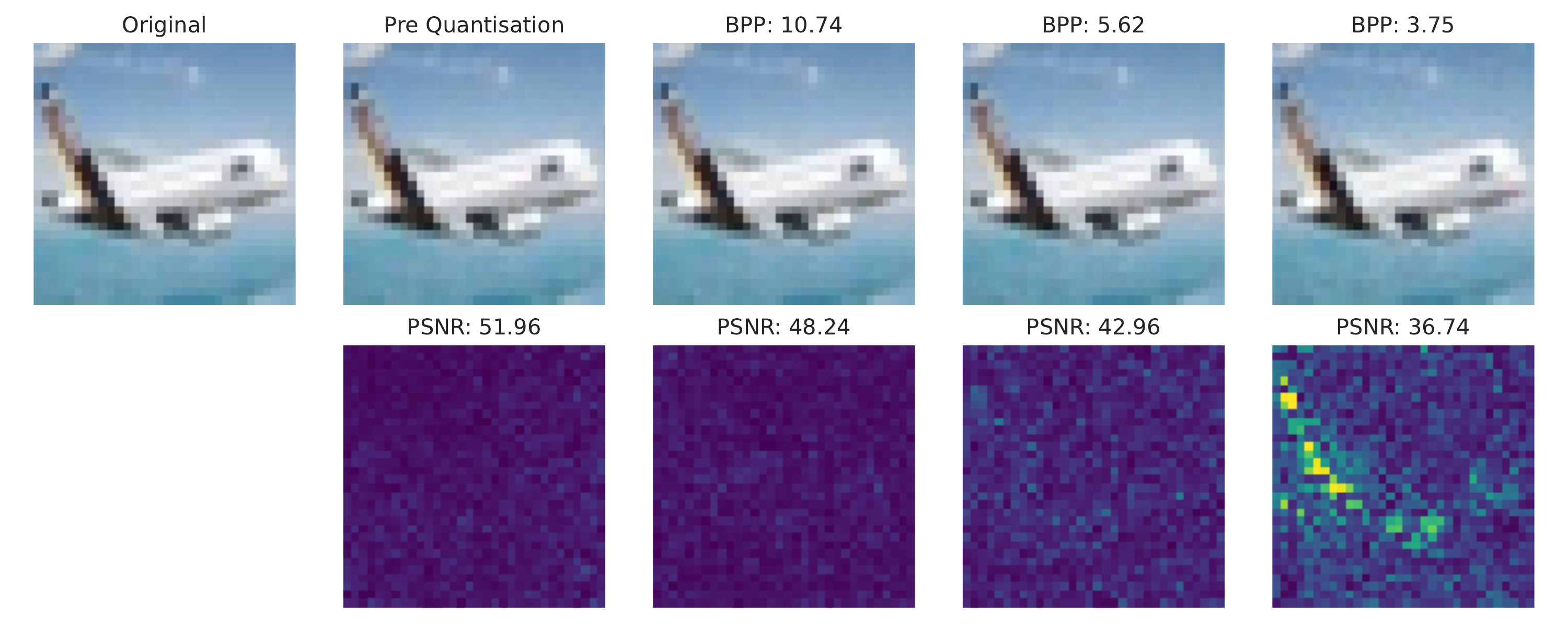}.
    \end{subfigure}
    \begin{subfigure}{0.49\linewidth}
        \centering
        \includegraphics[width=\textwidth]{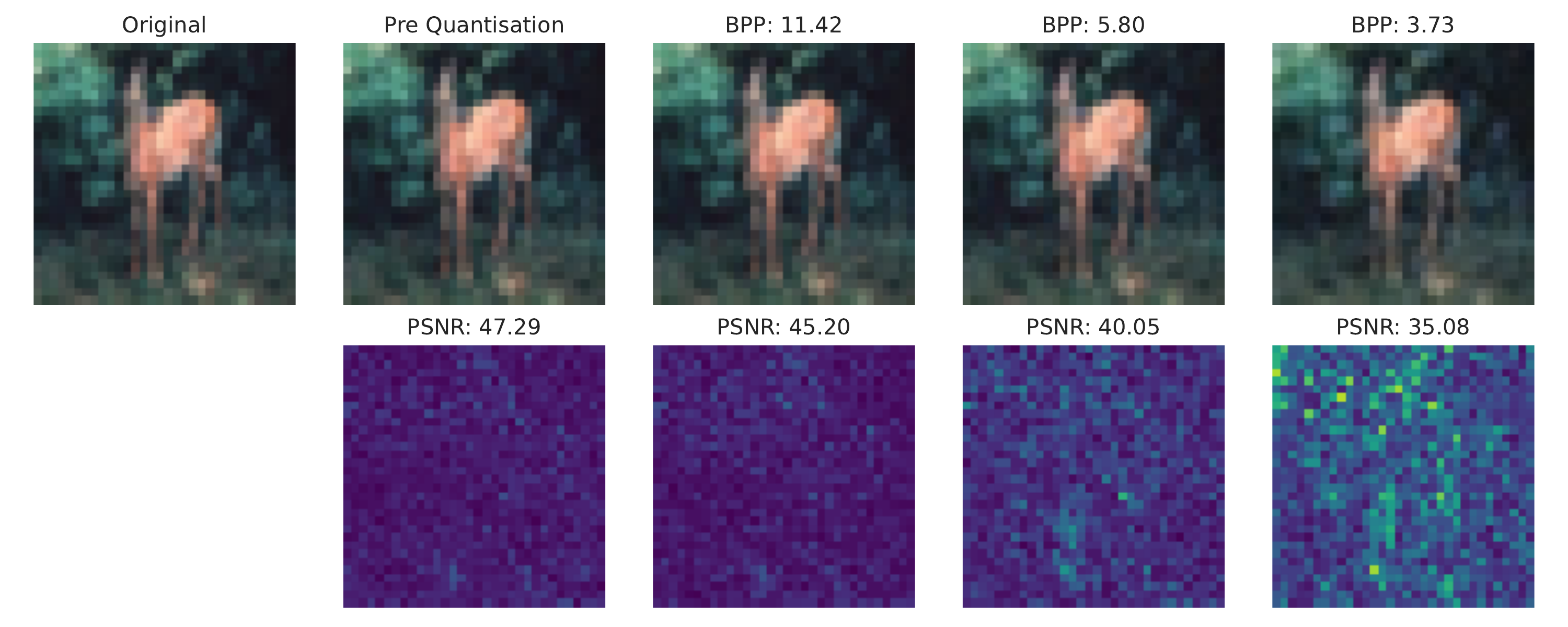}.
    \end{subfigure}
    \begin{subfigure}{0.49\linewidth}
        \centering
        \includegraphics[width=\textwidth]{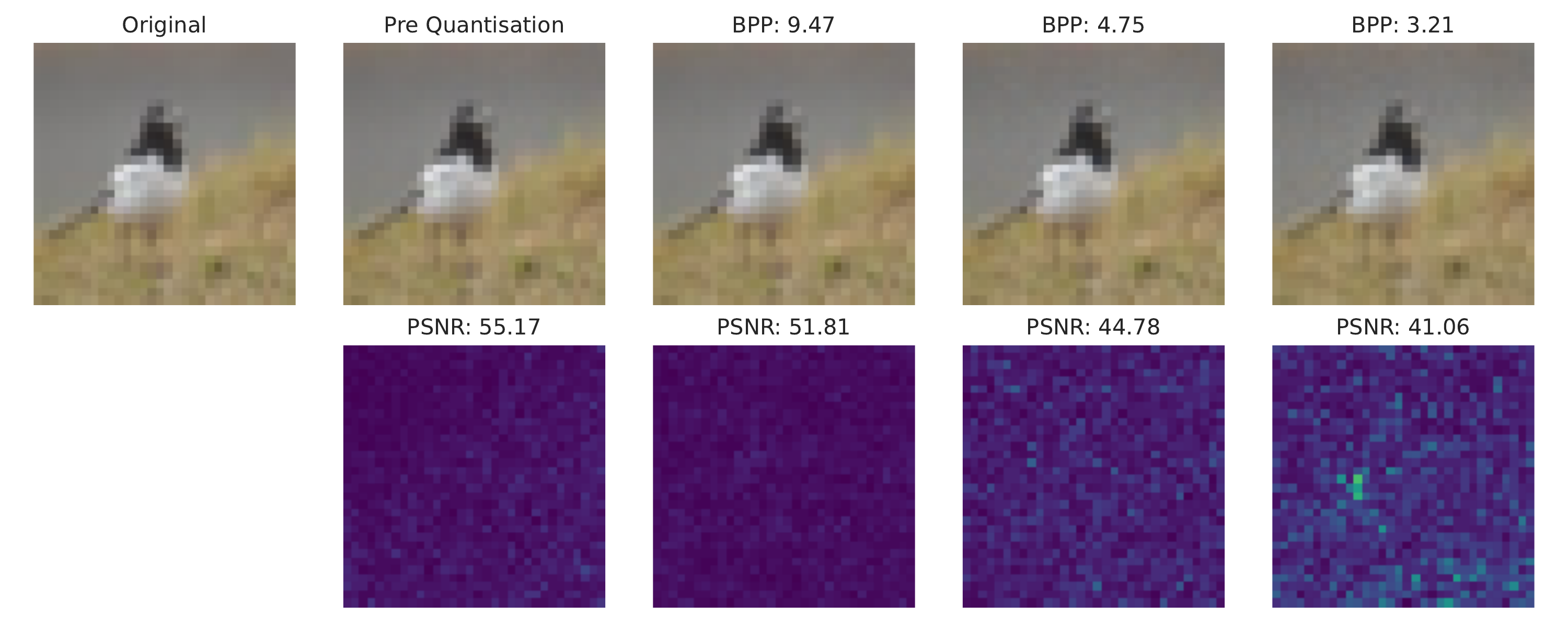}.
    \end{subfigure}
    \begin{subfigure}{0.49\linewidth}
        \centering
        \includegraphics[width=\textwidth]{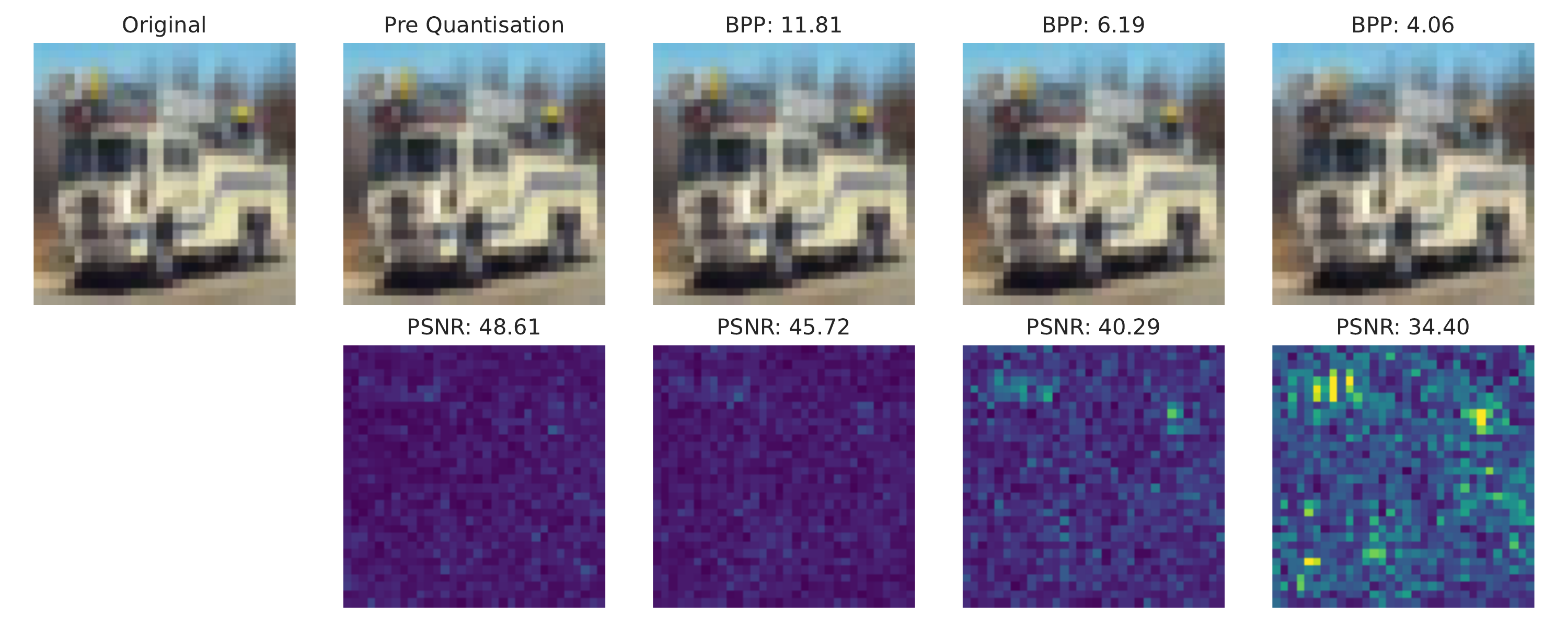}.
    \end{subfigure}
    \begin{subfigure}{0.49\linewidth}
        \centering
        \includegraphics[width=\textwidth]{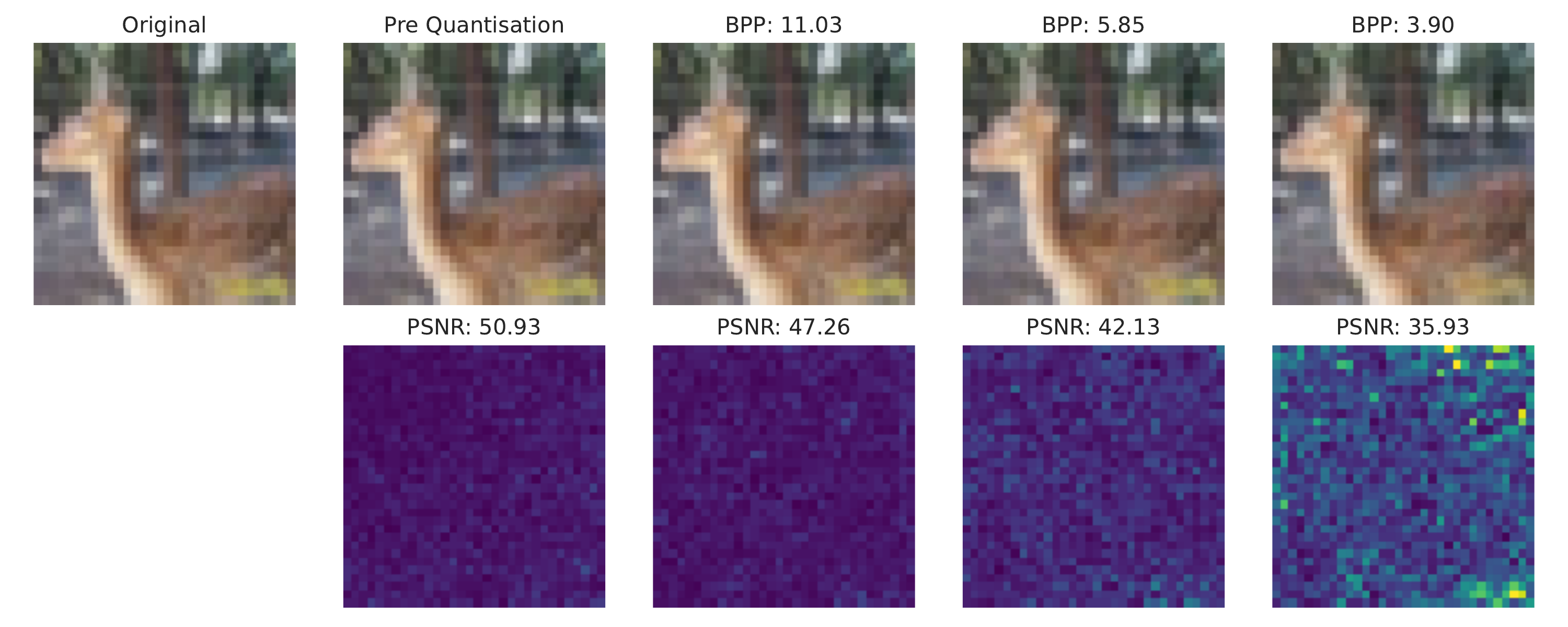}.
    \end{subfigure}
    \caption{More qualitative results from the Cifar10 dataset.}
    \label{fig:more_cifar10_qualitative}
\end{figure}

Figure \ref{fig:more_cifar10_qualitative} shows more qualitative results on Cifar10 for various rate/distortions trade-offs.

\begin{figure}[t]
\begin{subfigure}{0.32\textwidth}
\centering
    \includegraphics[width=\linewidth]{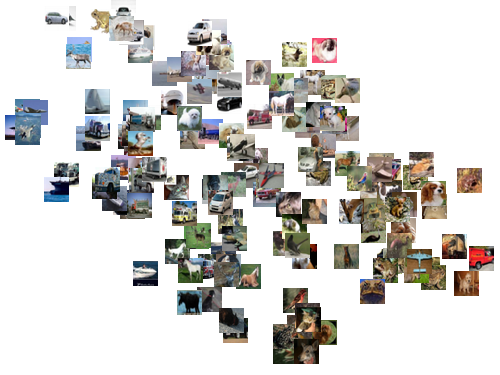}
    \caption{}
\end{subfigure}%
\begin{subfigure}{0.32\textwidth}
\centering
    \includegraphics[width=\linewidth]{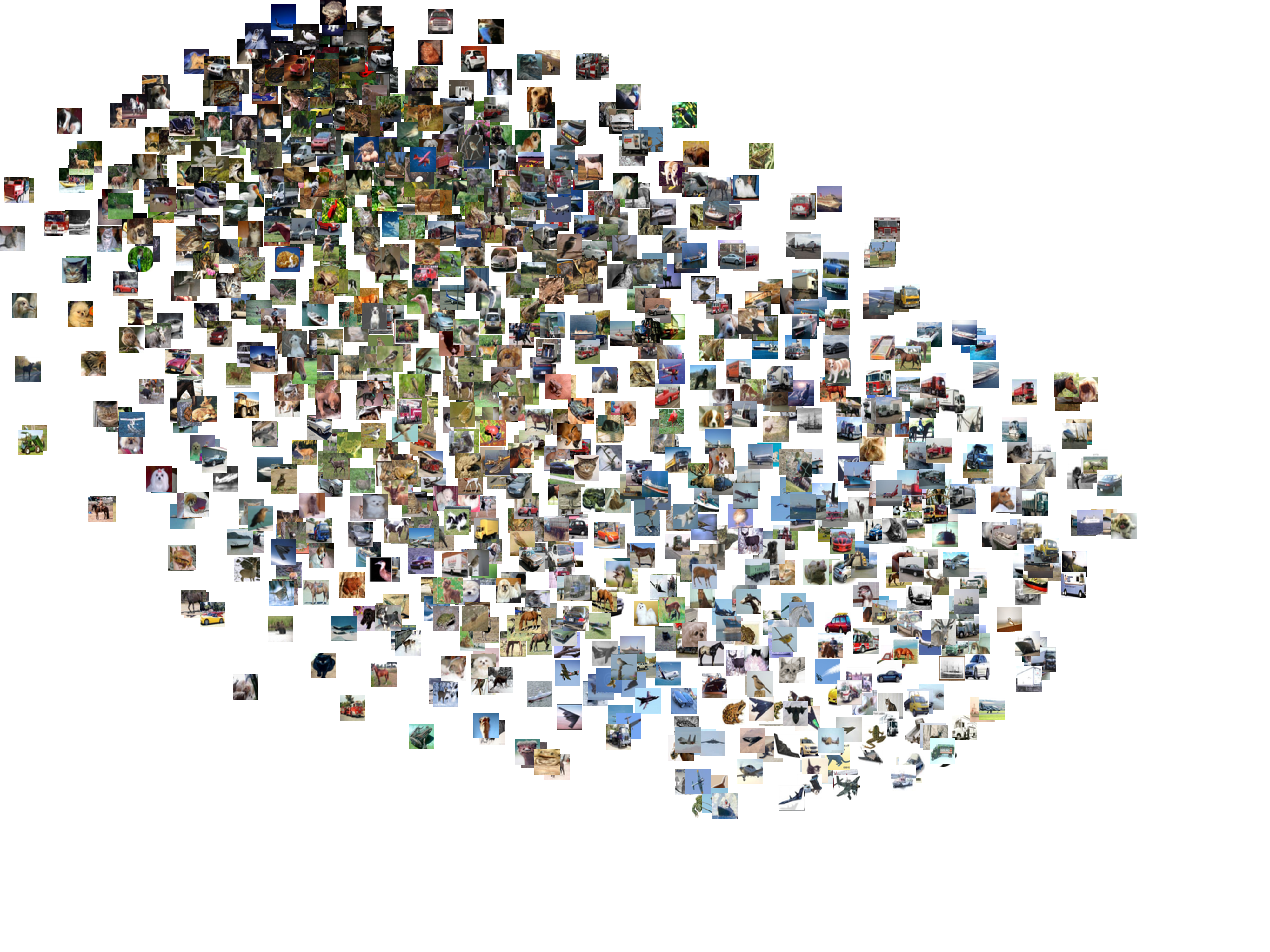}
    \caption{}
\end{subfigure}
\begin{subfigure}{0.35\textwidth}
\centering
    \includegraphics[width=\linewidth]{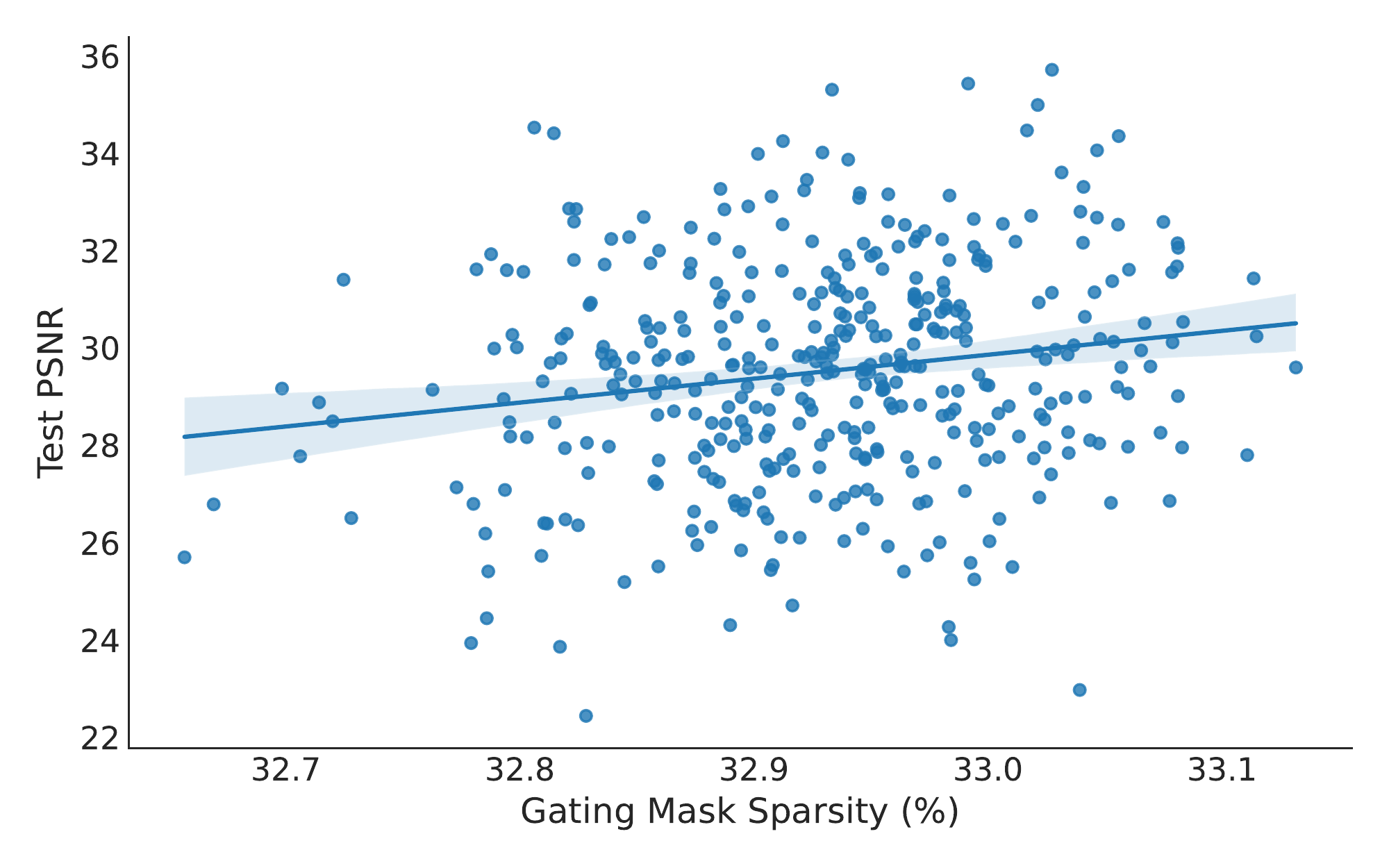}
    \caption{}
\end{subfigure}
\caption{t-SNE \citep{van2008visualizing} visualisation of gating masks $\rmG_{\mathtt{low}}$ after adaptation on CIFAR-10. (a) Full test set results (b) zoomed-in (c) Correlation between gating mask sparsity and performance level. Sparsity calculated as the fraction of sparse weights (i.e. $|(\rmG_{\mathtt{low}} \odot \rmW)_{ij}| < 0.001$) relative to the total number of all weights in the network.}
\label{fig:appendix_cifar10_gating}
\end{figure}

In addition, we provide a further analysis of gating masks using a similar t-SNE projection as shown in the main text for Cifar-10 as well as an analysis of sparsity level and reconstruction correlation in Figure \ref{fig:appendix_cifar10_gating}. With regards to correlation, it is firstly worth noting that there is little variation in the total sparsity level (reaching from 32.5 - 33.2). Secondly, we observe only very weak correlation (Pearson’s correlation coefficient: 0.177) suggesting that no straightforward relationship between sparsity and performance exists.

\subsection{Kodak}

\begin{figure}
    \centering
    \begin{subfigure}{0.85\linewidth}
        \centering
        \includegraphics[width=\textwidth]{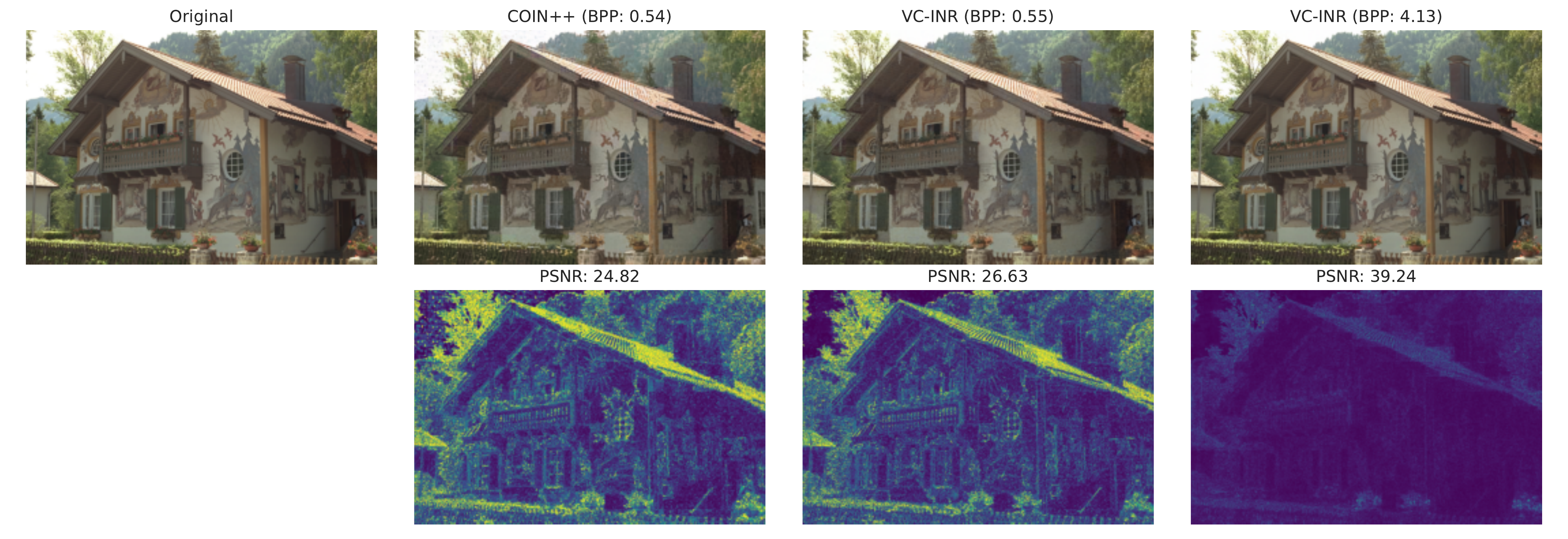}
        \caption{Compared with COIN++ \citep{dupont2022coinpp}.}
    \end{subfigure}
    \begin{subfigure}{0.85\linewidth}
        \centering
        \includegraphics[width=\textwidth]{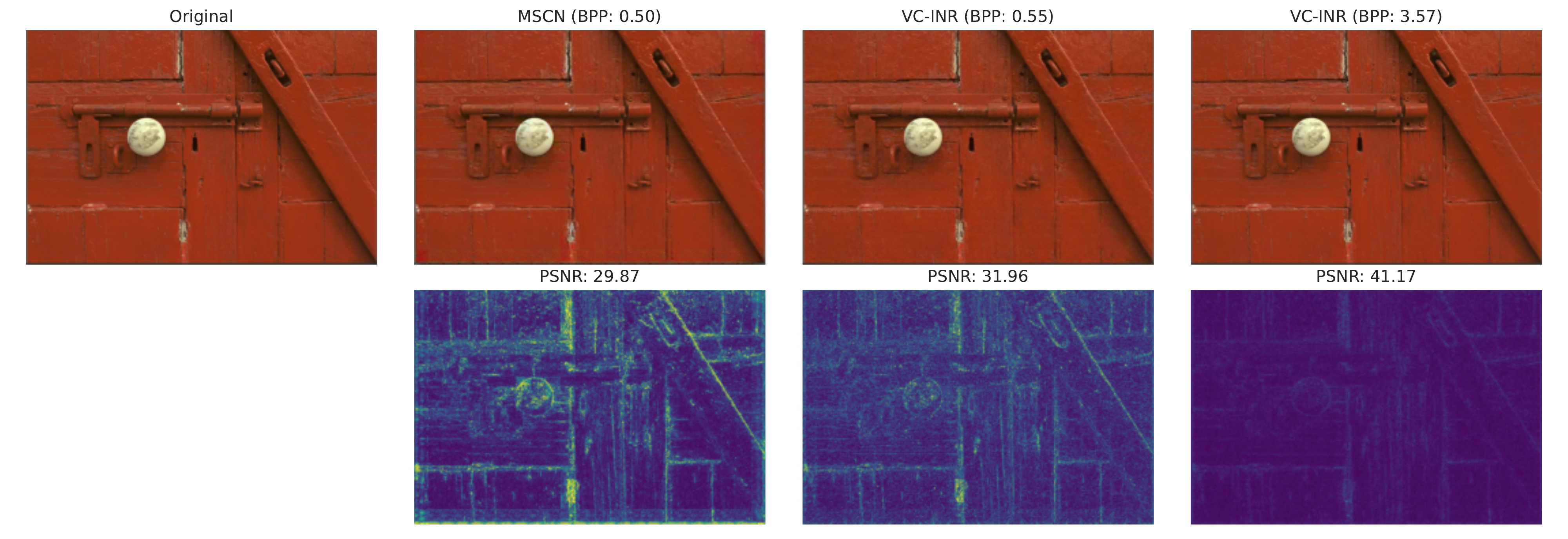}
        \caption{Compared with MSCN \citep{schwarz2022meta}.}
    \end{subfigure}
    \begin{subfigure}{0.85\linewidth}
        \centering
        \includegraphics[width=\textwidth]{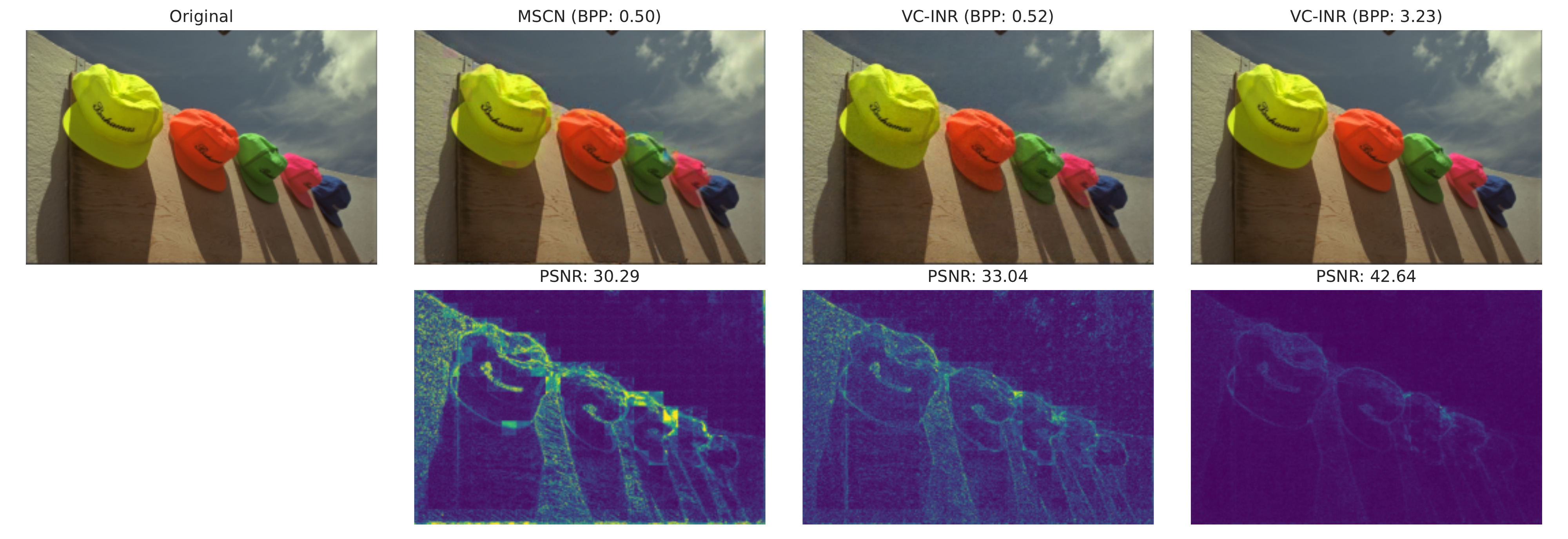}
        \caption{Compared with MSCN \citep{schwarz2022meta}.}
    \end{subfigure}
    \begin{subfigure}{0.85\linewidth}
        \centering
        \includegraphics[width=\textwidth]{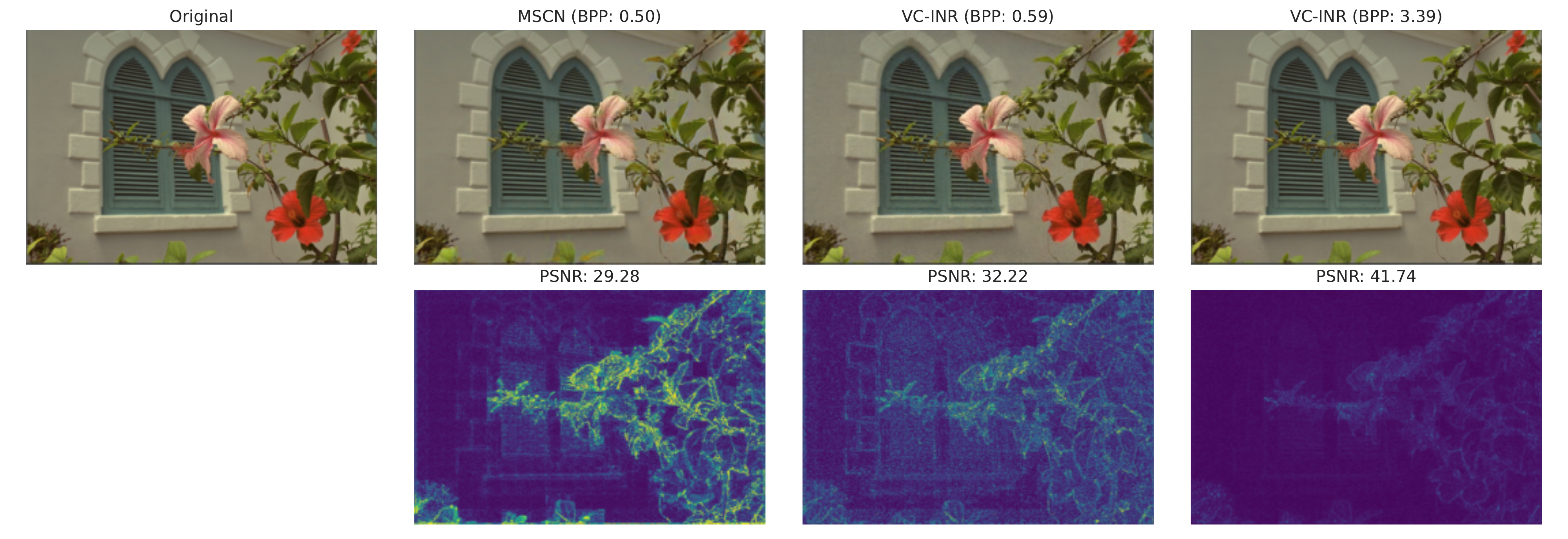}
        \caption{Compared with MSCN \citep{schwarz2022meta}.}
    \end{subfigure}
    %\begin{subfigure}{0.49\linewidth}
    %    \centering
    %    \includegraphics[width=\textwidth]{figures/kodak/vcinr_v_mscn_3.pdf}
    %    \caption{"}
    %\end{subfigure}
    %\begin{subfigure}{0.49\linewidth}
    %    \centering
    %    \includegraphics[width=\textwidth]{figures/kodak/vcinr_v_mscn_4.pdf}
    %    \caption{"}
    %\end{subfigure}
    %\begin{subfigure}{0.49\linewidth}
    %    \centering
    %    \includegraphics[width=\textwidth]{figures/kodak/vcinr_v_mscn_5.pdf}
    %    \caption{}
    %\end{subfigure}
    %\begin{subfigure}{0.49\linewidth}
    %    \centering
    %    \includegraphics[width=\textwidth]{figures/kodak/vcinr_v_mscn_6.pdf}
    %    \caption{}
    %\end{subfigure}
    \caption{More qualitative results from the Kodak dataset. Shown are VC-INR models in comparison with other INR-based techniques at similar bit-rates (3rd column) as well as a high-quality model (last column).}
    \label{fig:more_kodak_qualitative}
\end{figure}

Figure \ref{fig:more_kodak_qualitative} shows more qualitative results on Kodak in comparison with COIN++ \citep{dupont2022coinpp} and MSCN \citep{schwarz2022meta}.

\subsection{UCF-101}

\begin{figure}[h]
    \centering
    \begin{subfigure}{0.33\linewidth}
        \centering
        \includegraphics[width=\textwidth]{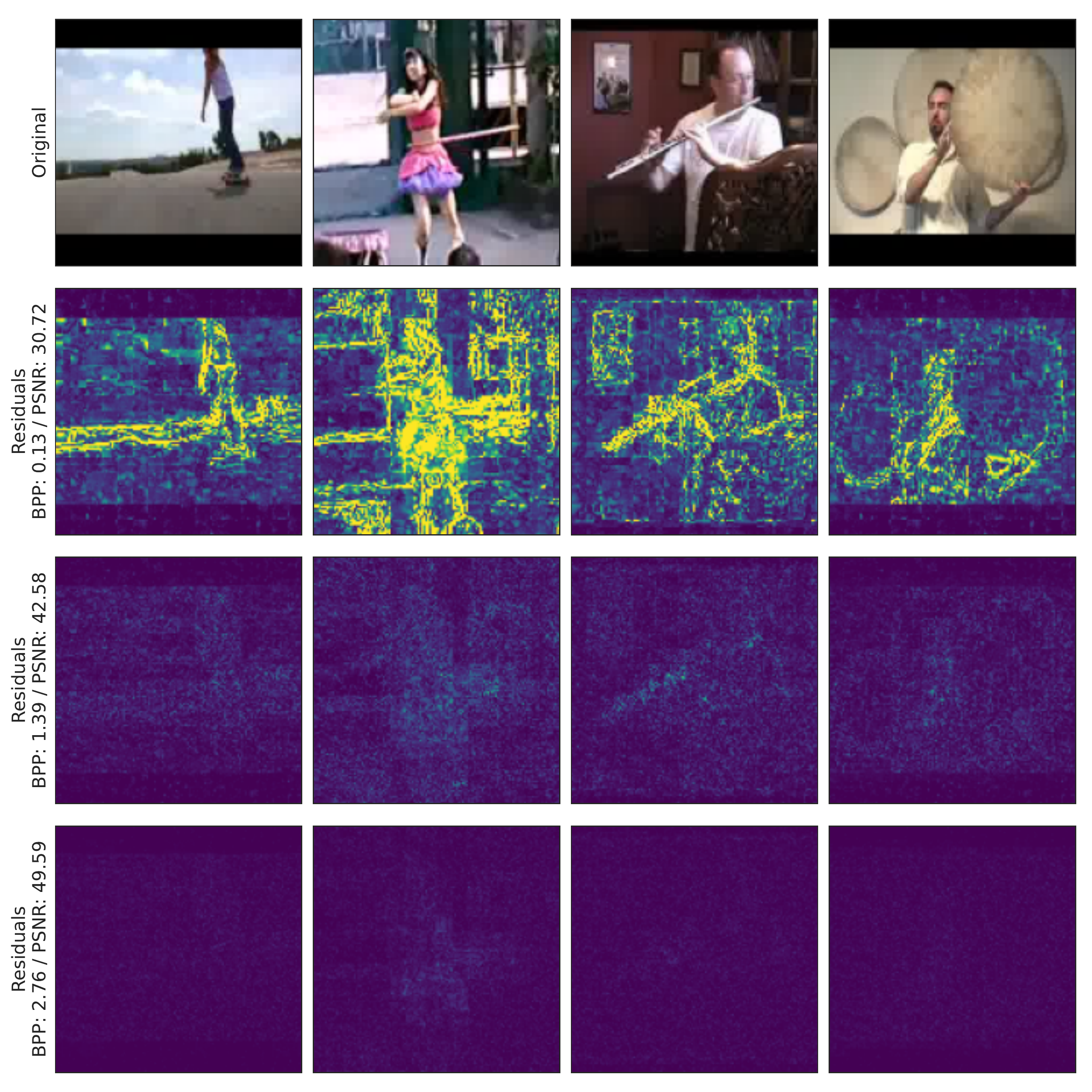}
        \caption{}
    \end{subfigure}
    \hfill
    \begin{subfigure}{0.33\linewidth}
        \centering
        \includegraphics[width=\textwidth]{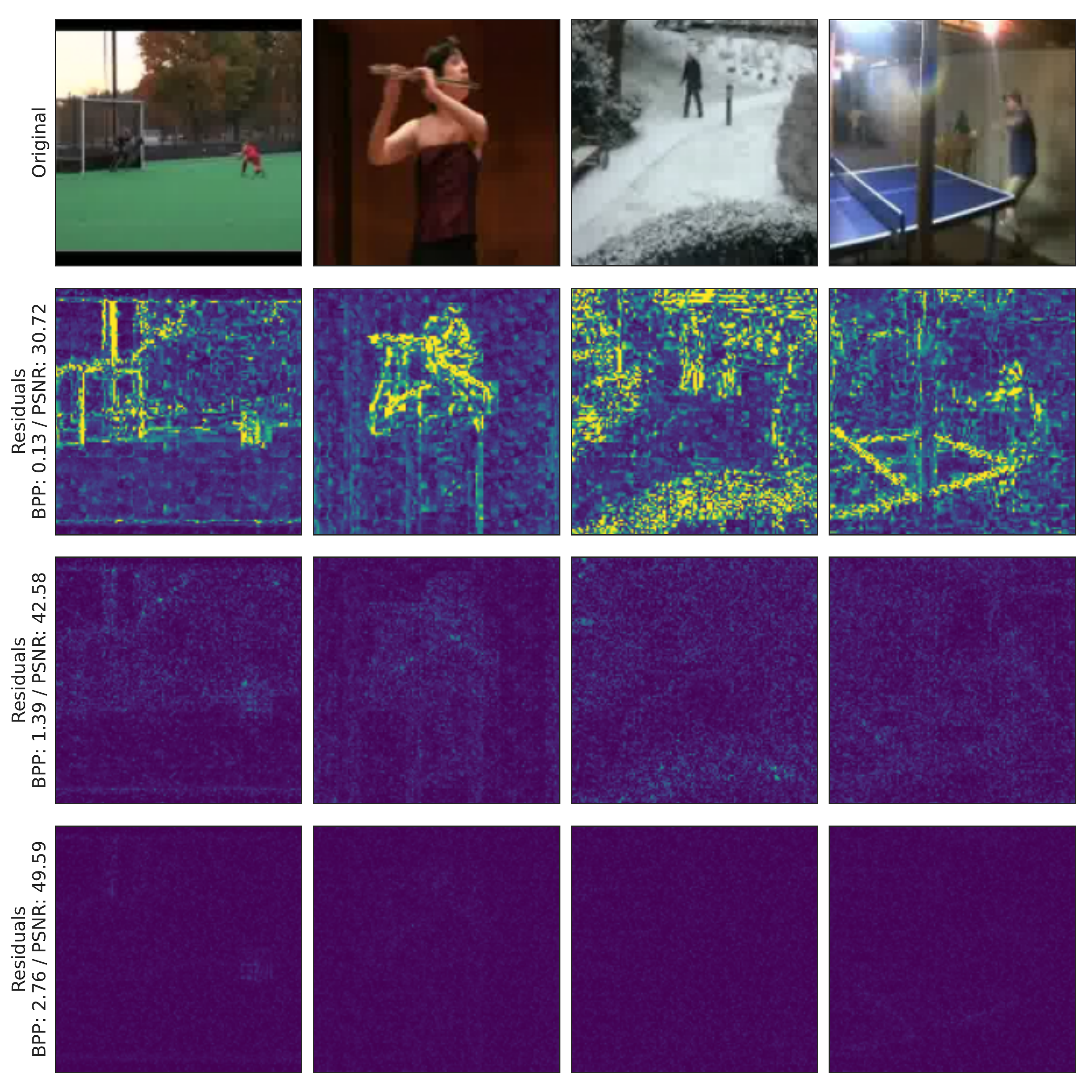}
        \caption{}
    \end{subfigure}
    \hfill
    \begin{subfigure}{0.33\linewidth}
        \centering
        \includegraphics[width=\textwidth]{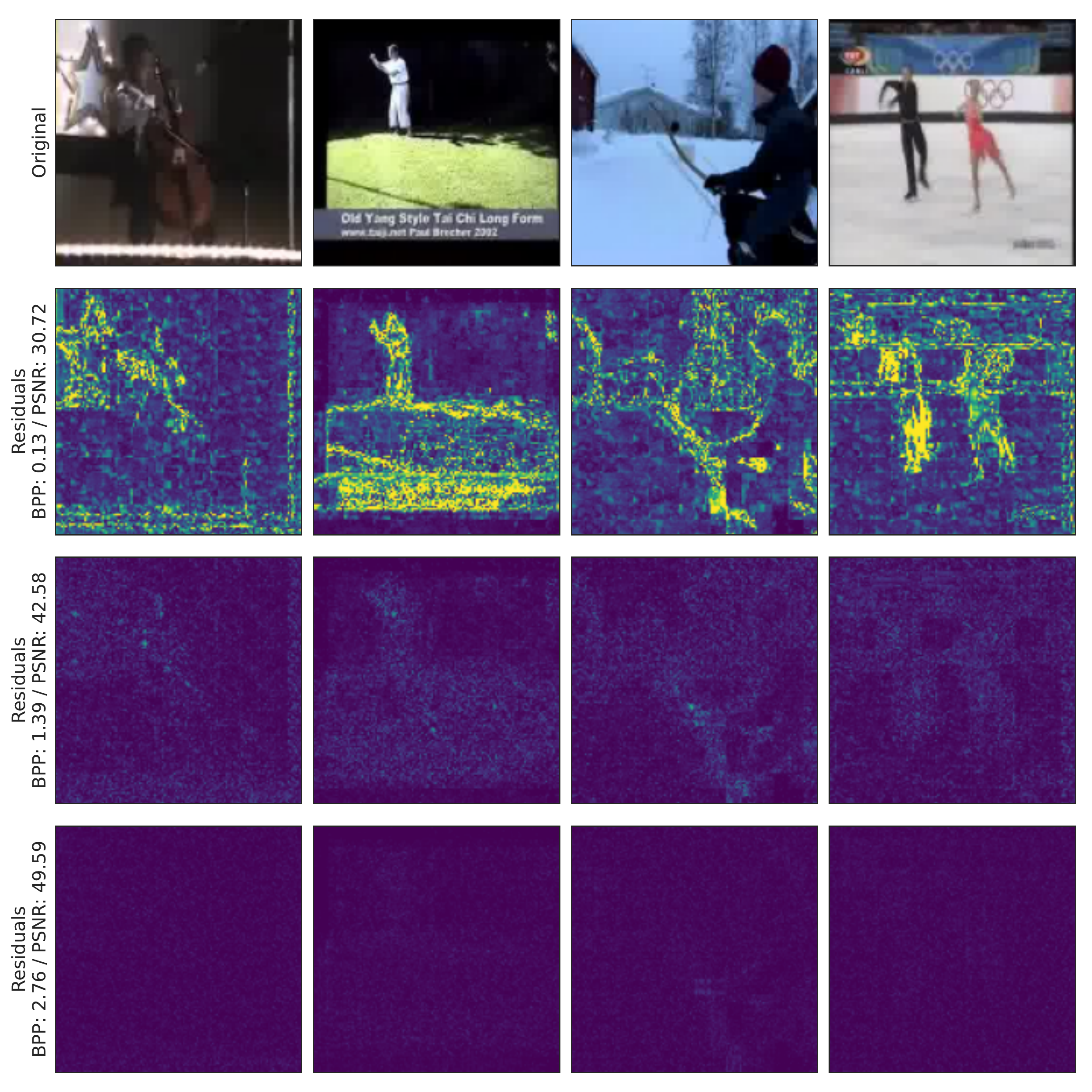}
        \caption{}
    \end{subfigure}

    \caption{More qualitative results from the UCF-101 dataset. Shown are VC-INR models at varying quality rates.}
    \label{fig:more_ucf_qualitative}
\end{figure}

Figure \ref{fig:more_ucf_qualitative} shows more qualitative results on frames from the UCF-101 dataset. We provide links to each of the reconstruction video clips and its residual in comparison with the original video in Table \ref{tab:more_ucf_qualitative}.

\begin{table}[]
\centering
\begin{tabular}{lcccc}
\hline
\footnotesize
%\multicolumn{1}{c}{} & \multicolumn{4}{c}{Quality}\\  
%\multicolumn{1}{c}{} & \multicolumn{1}{c}{Low (BPP: 0.13)} & \multicolumn{1}{c}{Medium (BPP: 1.39)} & \multicolumn{1}{c}{High (BPP: 2.76)} & \multicolumn{1}{c}{Best (BPP: 4.20)} \\ 

& \multicolumn{4}{c}{\textbf{Quality}} \\
& Low (BPP: 0.13) & Medium (BPP: 1.39) & High (BPP: 2.76) & Best (BPP: 4.20) \\ 

\hline

Example 1 & \href{https://drive.google.com/file/d/1bLqIWXBnxAv_6P0pgWXfFq2N1BXuSSGN/view?usp=sharing}{here} & \href{https://drive.google.com/file/d/17HxhopRHHLpmg6fSW3q5Wka6j5itWEp4/view?usp=sharing}{here} & \href{https://drive.google.com/file/d/1bkdHW4km8b8f_9fN4eQegTk2K8PQ1VNQ/view?usp=sharing}{here} & \href{https://drive.google.com/file/d/1zKt5rPduWh6biVjeIHxsSUf3ofUYtq5R/view?usp=sharing}{here} \\
Example 2 & \href{https://drive.google.com/file/d/1Z6RgEZjeBALsLE0UxZ0Xz6t0VsNKAvrV/view?usp=sharing}{here} & \href{https://drive.google.com/file/d/1-EWmFK5qw4FWLb4KpERSzFv56vr7tCbR/view?usp=sharing}{here} & \href{https://drive.google.com/file/d/1mnAZWTG6wE_p1-9j8PRN9I7uIrOodCWn/view?usp=sharing}{here} & \href{https://drive.google.com/file/d/18a8IKEXkK0HNSElPapwsEsbd3tWkzr24/view?usp=sharing}{here} \\
Example 3 & \href{https://drive.google.com/file/d/1-yVDu4KSoBg0OgZcqczoOUXFbeerfB4b/view?usp=sharing}{here} & \href{https://drive.google.com/file/d/1Sm1Ucm4tTXUrD-jGXHtRHskVrcsv9fet/view?usp=sharing}{here} & \href{https://drive.google.com/file/d/1jg6769bUpyeQJVX9_gb3o0ySKtfT8Hwv/view?usp=sharing}{here} & \href{https://drive.google.com/file/d/1ZN39bLJWlw-wbSt671LJkf2iigX0eMQg/view?usp=sharing}{here} \\
Example 4 & \href{https://drive.google.com/file/d/1-xoMc99qSk-qNnt3iZVl-lwRB38PmUPL/view?usp=sharing}{here} & \href{https://drive.google.com/file/d/1GTQl10oM-gGs5hQZeIzeslqZkF53DD2U/view?usp=sharing}{here} & \href{https://drive.google.com/file/d/16o0TiZSF3LHB3KC2QrLkI7I6VXhE9nqn/view?usp=sharing}{here} & \href{https://drive.google.com/file/d/1wCTA-hwHBOO80WcfMjQUiZRyve-_cfb_/view?usp=sharing}{here} \\
Example 5 & \href{https://drive.google.com/file/d/13h6SoiFYgK8kgBMjqYCyM-ICZ1JqM4FM/view?usp=sharing}{here} & \href{https://drive.google.com/file/d/1Yl8qlXRsBqcwDL-7gilUMohDKCyrUxgi/view?usp=sharing}{here} & \href{https://drive.google.com/file/d/1S9cFmL2pQepSxkK_GfGqIKwux4dT_Ex6/view?usp=sharing}{here} & \href{https://drive.google.com/file/d/1cYkRp826WGYvHCSzQMLKwGYeftgiAOS4/view?usp=sharing}{here} \\
Example 6 & \href{https://drive.google.com/file/d/1g9tabwNMNVQm1jn2oACbcWR_J4VggJmW/view?usp=sharing}{here} & \href{https://drive.google.com/file/d/1ZXKUfK_wLG8J_T8mRecEaZjY8lfduSxs/view?usp=sharing}{here} & \href{https://drive.google.com/file/d/1RvmnRF6-_g_hMDpQlcrePHrh2cs14beW/view?usp=sharing}{here} & \href{https://drive.google.com/file/d/1rlTcTSRiOrUNg7eXZooM0kO2Dtm7BDZU/view?usp=sharing}{here} \\
Example 7 & \href{https://drive.google.com/file/d/1FcY8QEtXFJYTU1qYA71d6moduCtiZgNP/view?usp=sharing}{here} & \href{https://drive.google.com/file/d/17KrhCZUGDQUZVCLlif3E3RWEaWCROWsu/view?usp=sharing}{here} & \href{https://drive.google.com/file/d/168PIeky6XSHDiF3eER0Jy3me9Xovi2dN/view?usp=sharing}{here} & \href{https://drive.google.com/file/d/1DAuS8PK8KMs1WtgT15UIw8b5RA2BIWfe/view?usp=sharing}{here} \\
Example 8 & \href{https://drive.google.com/file/d/1HjrpXD2ifE8goXVLtM2kZ65bsCTG94NV/view?usp=sharing}{here} & \href{https://drive.google.com/file/d/1hwO9ktTVSI4-lGAZBUX5YvNE9c-tl4cJ/view?usp=sharing}{here} & \href{https://drive.google.com/file/d/1J1ksjV315SnsijHYxXXAjFH1o8ZFRURI/view?usp=sharing}{here} & \href{https://drive.google.com/file/d/16cGKSPcIIJ0mhiGzNo7OeLd4B0BSUebh/view?usp=sharing}{here} \\
Example 9 & \href{https://drive.google.com/file/d/12I5HEcRUxFItdq6nC9h4jwgqtNXuIEl5/view?usp=sharing}{here} & \href{https://drive.google.com/file/d/1L6XIsF1vo9AET1qVhpRsuxf7mCnjvj-4/view?usp=sharing}{here} & \href{https://drive.google.com/file/d/1o53TGUklPFtKdcf1GcmoqlcrTDPsWdPS/view?usp=sharing}{here} & \href{https://drive.google.com/file/d/1wclmlGKDgBD48Y99CYgwHfiyQjDctEiG/view?usp=sharing}{here} \\
Example 10 &\href{https://drive.google.com/file/d/13ODPF5XQAzkiNkDG-WnJsVAMYMIYx3oq/view?usp=sharing}{here} &  \href{https://drive.google.com/file/d/1PMVLop8j7O1m1i_yhWGXRpzF0mYCrjLB/view?usp=sharing}{here} & \href{https://drive.google.com/file/d/1YDmxqGOcABHhNmHvop8eyz_k67_uc2nh/view?usp=sharing}{here} &\href{https://drive.google.com/file/d/1UIP3b-nPKuwWwkemsoOr4gaBy9D3gTJC/view?usp=sharing}{here} \\
Example 11 &\href{https://drive.google.com/file/d/1AYPVgYNGXKWUFLSy-Ib3jssCQo8eRv6E/view?usp=sharing}{here} &  \href{https://drive.google.com/file/d/1go0X5mIehp2AjPvN8W9gQhh6k88ElhBW/view?usp=sharing}{here} & \href{https://drive.google.com/file/d/16x-hnfhVehFiauEv69KZcx7tIEMnNTEG/view?usp=sharing}{here} &\href{https://drive.google.com/file/d/10LPAELNyJaMO10YX4hQ8COjhyUQL7V5T/view?usp=sharing}{here} \\
Example 12 &\href{https://drive.google.com/file/d/1bjrMqhont5CqWHLnE9eMpcCqsF7wgJWk/view?usp=sharing}{here} &  \href{ttps://drive.google.com/file/d/1pOWCnwBq_-jaoosDj3fuLw3Ch-btuVEC/view?usp=sharing}{here} & \href{https://drive.google.com/file/d/1CNz3EffQn9QdsAnxLOC8fTFPX0ZDxF32/view?usp=sharing}{here} & \href{https://drive.google.com/file/d/1m6tYbWud_2ekmo3ZMZMM5208yYLyJfdm/view?usp=sharing}{here} \\
Example 13 &\href{https://drive.google.com/file/d/1bOJJi5tQY7uqDTJUNSV1To3mhq6NkZxL/view?usp=sharing}{here} &  \href{https://drive.google.com/file/d/18KCbyulvn9wBrT0YsPWd7Pe1V_B9a5_V/view?usp=sharing}{here} & \href{https://drive.google.com/file/d/1C79OfnYt_iTPNahDREPfftFdtRXA0Beb/view?usp=sharing}{here} &\href{https://drive.google.com/file/d/1u5bV6b6lYCF4GP4Gpd6Tll0rMECL2UdZ/view?usp=sharing}{here} \\
Example 14 &\href{https://drive.google.com/file/d/1MP207R5TL64KUrs6cYuyFhK6DWJHBVik/view?usp=sharing}{here} &  \href{https://drive.google.com/file/d/1PiUC7tx8w_N31Io325K6ywrOn9SoQsxP/view?usp=sharing}{here} & \href{https://drive.google.com/file/d/1K1A73OnVnGfCO8DqoANBe8awjLKRtomX/view?usp=sharing}{here} &\href{https://drive.google.com/file/d/1ez6l8OnledBh0Xfn0sHK94ko2r5uIhmb/view?usp=sharing}{here} \\
Example 15 &\href{https://drive.google.com/file/d/10GgcN9dI13Q-YYeNavu1eTi0MU8RTNr4/view?usp=sharing}{here} &  \href{https://drive.google.com/file/d/1Rgjftl1lvuAy8rsj34gkwfaSebOBpFXq/view?usp=sharing}{here} & \href{https://drive.google.com/file/d/1m8OxvkEKtIKJKMaPlTQqa_UQdbGZ7eWY/view?usp=sharing}{here} &\href{https://drive.google.com/file/d/1GDXGFQbctAKJDbFGj99RlVgToIjAfNrF/view?usp=sharing}{here} \\
Example 16 &\href{https://drive.google.com/file/d/1LbaBOwWcpeDhD8JywxRacSFQ85x8_TQp/view?usp=sharing}{here} &  \href{https://drive.google.com/file/d/1lg2z569hVbEhwDRAKjI3XEfaX4UK7o2n/view?usp=sharing}{here} & \href{https://drive.google.com/file/d/1FaP7VC6-H3l21wPyXalfTTvKzw_ssiow/view?usp=sharing}{here} &\href{https://drive.google.com/file/d/1UDhOlgEKRH054uZqeSKZ_6jgPcIJyajB/view?usp=sharing}{here} \\
\hline

\end{tabular}

\caption{More qualitative examples from the UCF-101 datasets. Shown are full video reconstructions and residuals of various VC-INR models at varying bit-rates.}
\label{tab:more_ucf_qualitative}
\end{table}
\section{Hyperparameters}
\subsection{Compression experiments}

We show hyperparameters for both INR training and subsequent compression training for CIFAR-10 in Table \ref{tab:compress_cifar_hps}, for Kodak in Table \ref{tab:compress_kodak_hps}, for ERA5 in Table \ref{tab:compress_era5_hps}, for LibriSpeech in Table \ref{tab:compress_libri_hps} and for UCF-101 in Table \ref{tab:compress_ucf_hps}.
\begin{table}[h]
\footnotesize
\centering
\caption{Hyperparameters for compression experiments on CIFAR-10.}
\label{tab:compress_cifar_hps}
\begin{tabular}{llc}
\toprule
\textbf{Parameter} & \textbf{Considered range} & \textbf{Comment} \\ 
\toprule
\textbf{INR training} & & \\
Patching & \{False\} & \\
Activation function & $\{h(x): \sin(\omega_0x) \text{ (SIREN)}\}$ & \\
$\omega_0$ & \{30\} & \\
Network depth & $\{15\}$ & \\
Network width & $\{512\}$ & \\

Batch size per device & $\{32, 64\}$ & \\
Num devices & $\{8\}$ & \\
Optimiser & \{Adam\} & \\
Outer learning rate & \{$3\cdot 10^{-6}$\} & \\
Num inner steps & \{3\} & \\
Meta-learn $\boldsymbol\phi$ init. & \{True\} & \\
Meta SGD range & \{[-5.0, 5.0]\} & (Max./Min. for Meta-SGD LRs)\\
Meta SGD init range & \{[1.0, 1.0]\} & (Uniformly sampled).\\

\midrule
\textbf{$\boldsymbol\phi \rightarrow \{\rmG_{\mathtt{low}}^{(1)}, \dots, \rmG_{\mathtt{low}}^{(L)}\}$ network} & & \\
\midrule

$\dim(\boldsymbol\phi)$ & \{2048, 3072, 4096\} &  \\
Use LayerNorm & \{True\} &  \\
Network width & \{6144\} &  \\
Residual blocks & $\{2\}$ & \\
Activation function & \{Leaky Relu\} & \citep{xu2015empirical}\\
Adapt first Layer & \{False\} & Apply low-rank gating to 1st layer?\\

\midrule
\textbf{Quantiser training} & & \\
\midrule
Normalise $\boldsymbol\phi$ & $\{\text{True}\}$ & Per dim. $\frac{\boldsymbol\phi_i - \hat{\boldsymbol\mu}_i}{\hat{\boldsymbol\sigma}_i}$ based on $\boldsymbol\phi$ train-set stats.\\
$\lambda$ ($\mathcal{L}_{\mathtt{distortion}}$ penalty) & \{0.33, 0.66, 1.0, 3.33, 6.66\} & \\
Analysis transform ($g_a$) Residual blocks & $\{1\}$ & \\
$g_a$ Network width & $\{2048, 4096, 5120\}$ & \\
$g_a$ Activation function & SeLU & \citep{klambauer2017self} \\
$\dim(\rvy)$ & \{1024, 2048, 4096, 5120\} &  \\

Synthesis transform $g_s$  & Same as $g_a$ &  \\
Optimiser & \{Adam\} & \\
Learning rate & \{$1\cdot 10^{-4}$\} & \\
Batch size per device & $\{32, 64\}$ & \\
Num devices & $\{1\}$ & \\

\bottomrule
\end{tabular}
\end{table}
\begin{table}[h]
\footnotesize
\centering
\caption{Hyperparameters for compression experiments on Div2k/Kodak.}
\label{tab:compress_kodak_hps}
\begin{tabular}{llc}
\toprule
\textbf{Parameter} & \textbf{Considered range} & \textbf{Comment} \\ 
\toprule
\textbf{INR training} & & \\
Pre-training on & \{Div2k\} & \tiny as in 
\citep{schwarz2022meta, strumpler2022implicit}\\
Patching & \{$(32 \times 32)$\} & Dividing $768\times512$ images.\\
Activation function & $\{h(x): \sin(\omega_0x) \text{ (SIREN)}\}$ & \\
$\omega_0$ & \{30\} & \\
Network depth & $\{15\}$ & \\
Network width & $\{512\}$ & \\

Batch size per device & $\{32\}$ & \\
Num devices & $\{8\}$ & \\
Optimiser & \{Adam\} & \\
Outer learning rate & \{$3\cdot 10^{-6}$\} & \\
Num inner steps & \{3\} & \\
Meta-learn $\boldsymbol\phi$ init. & \{True\} & \\
Meta SGD range & \{[-5.0, 5.0]\} & (Max./Min. for Meta-SGD LRs)\\
Meta SGD init range & \{[1.0, 1.0]\} & (Uniformly sampled).\\

\midrule
\textbf{$\boldsymbol\phi \rightarrow \{\rmG_{\mathtt{low}}^{(1)}, \dots, \rmG_{\mathtt{low}}^{(L)}\}$ network} & & \\
\midrule

$\dim(\boldsymbol\phi)$ & \{512, 1024\} &  \\
Use LayerNorm & \{True\} &  \\
Network width & \{4096\} &  \\
Residual blocks & $\{1\}$ & \\
Activation function & \{Leaky Relu\} & \citep{xu2015empirical}\\
Adapt first Layer & \{False\} & Apply low-rank gating to 1st layer?\\

\midrule
\textbf{Quantiser training} & & \\
\midrule
Normalise $\boldsymbol\phi$ & $\{\text{True}\}$ & Per dim. $\frac{\boldsymbol\phi_i - \hat{\boldsymbol\mu}_i}{\hat{\boldsymbol\sigma}_i}$ based on $\boldsymbol\phi$ train-set stats.\\
$\lambda$ ($\mathcal{L}_{\mathtt{distortion}}$ penalty) & \{0.01, 0.033, 0.1, 0.33, 0.66, 1.0\} & \\
Analysis transform ($g_a$) Residual blocks & $\{1\}$ & \\
$g_a$ Network width & $\{256, 512, 1024\}$ & \\
$g_a$ Activation function & SeLU & \citep{klambauer2017self} \\
$\dim(\rvy)$ & $\{256, 512, 1024\}$ &  \\

Synthesis transform $g_s$  & Same as $g_a$ &  \\
Optimiser & \{Adam\} & \\
Learning rate & \{$1\cdot 10^{-4}$\} & \\
Batch size per device & $\{128\}$ & \\
Num devices & $\{1\}$ & \\

\bottomrule
\end{tabular}
\end{table}
\begin{table}[h]
\footnotesize
\centering
\caption{Hyperparameters for compression experiments on ERA5 ($16\times$).}
\label{tab:compress_era5_hps}
\begin{tabular}{llc}
\toprule
\textbf{Parameter} & \textbf{Considered range} & \textbf{Comment} \\ 
\toprule
\textbf{INR training} & & \\
Patching & \{False\} & \\
Activation function & $\{h(x): \sin(\omega_0x) \text{ (SIREN)}\}$ & \\
$\omega_0$ & \{30\} & \\
Network depth & $\{10\}$ & \\
Network width & $\{384\}$ & \\

Batch size per device & $\{4\}$ & \\
Num devices & $\{4\}$ & \\
Optimiser & \{Adam\} & \\
Outer learning rate & \{$3\cdot 10^{-6}$\} & \\
Num inner steps & \{3\} & \\
Meta-learn $\boldsymbol\phi$ init. & \{True\} & \\
Meta SGD range & \{[-5.0, 5.0]\} & (Max./Min. for Meta-SGD LRs)\\
Meta SGD init range & \{[1.0, 1.0]\} & (Uniformly sampled).\\

\midrule
\textbf{$\boldsymbol\phi \rightarrow \{\rmG_{\mathtt{low}}^{(1)}, \dots, \rmG_{\mathtt{low}}^{(L)}\}$ network} & & \\
\midrule

$\dim(\boldsymbol\phi)$ & \{4, 8, 12, 32, 64, 128\} &  \\
Use LayerNorm & \{True\} &  \\
Network width & \{512\} &  \\
Residual blocks & $\{2\}$ & \\
Activation function & \{Leaky Relu\} & \citep{xu2015empirical}\\
Adapt first Layer & \{False\} & Apply low-rank gating to 1st layer?\\

\midrule
\textbf{Quantiser training} & & \\
\midrule
Normalise $\boldsymbol\phi$ & $\{\text{True}\}$ & Per dim. $\frac{\boldsymbol\phi_i - \hat{\boldsymbol\mu}_i}{\hat{\boldsymbol\sigma}_i}$ based on $\boldsymbol\phi$ train-set stats.\\
$\lambda$ ($\mathcal{L}_{\mathtt{distortion}}$ penalty) & \{0.001, 0.01, 0.01, 0.1\} & \\
Analysis transform ($g_a$) Residual blocks & $\{2\}$ & \\
$g_a$ Network width & $\{8, 12, 32, 64, 128\}$ & \\
$g_a$ Activation function & SeLU & \citep{klambauer2017self} \\
$\dim(\rvy)$ & \{8, 12, 32, 64, 128\} &  \\

Synthesis transform $g_s$ & Same as $g_a$ &  \\
Optimiser & \{Adam\} & \\
Learning rate & \{$1\cdot 10^{-4}$\} & \\
Batch size per device & $\{128, 256\}$ & \\
Num devices & $\{1\}$ & \\

\bottomrule
\end{tabular}
\end{table}
\begin{table}[h]
\footnotesize
\centering
\caption{Hyperparameters for compression experiments on LibriSpeech.}
\label{tab:compress_libri_hps}
\begin{tabular}{llc}
\toprule
\textbf{Parameter} & \textbf{Considered range} & \textbf{Comment} \\ 
\toprule
\textbf{INR training} & & \\
Patching & \{$(200, 400, 800)$\} & Dividing $48$k dim. audio signal.\\
Activation function & $\{h(x): \sin(\omega_0x) \text{ (SIREN)}\}$ & \\
$\omega_0$ & \{10, 30, 50\} & \\
Network depth & $\{10\}$ & \\
Network width & $\{512\}$ & \\

Batch size per device & $\{32, 64\}$ & \\
Num devices & $\{1\}$ & \\
Optimiser & \{Adam\} & \\
Outer learning rate & \{$3\cdot 10^{-6}$\} & \\
Num inner steps & \{3\} & \\
Meta-learn $\boldsymbol\phi$ init. & \{True\} & \\
Meta SGD range & \{[-5.0, 5.0]\} & (Max./Min. for Meta-SGD LRs)\\
Meta SGD init range & \{[1.0, 1.0]\} & (Uniformly sampled).\\

\midrule
\textbf{$\boldsymbol\phi \rightarrow \{\rmG_{\mathtt{low}}^{(1)}, \dots, \rmG_{\mathtt{low}}^{(L)}\}$ network} & & \\
\midrule

$\dim(\boldsymbol\phi)$ & \{64, 128, 256, 512, 1024\} &  \\
Use LayerNorm & \{True\} &  \\
Network width & \{512, 512, 768, 1536, 3072\} &  \\
Residual blocks & $\{2\}$ & \\
Activation function & \{Leaky Relu\} & \citep{xu2015empirical}\\
Adapt first Layer & \{False\} & Apply low-rank gating to 1st layer?\\

\midrule
\textbf{Quantiser training} & & \\
\midrule
Normalise $\boldsymbol\phi$ & $\{\text{True}\}$ & Per dim. $\frac{\boldsymbol\phi_i - \hat{\boldsymbol\mu}_i}{\hat{\boldsymbol\sigma}_i}$ based on $\boldsymbol\phi$ train-set stats.\\
$\lambda$ ($\mathcal{L}_{\mathtt{distortion}}$ penalty) & \{1.0, 10.0, 100.0\} & \\
Analysis transform ($g_a$) Residual blocks & $\{2\}$ & \\
$g_a$ Network width & $\{128, 256, 512, 1024\}$ & \\
$g_a$ Activation function & SeLU & \citep{klambauer2017self} \\
$\dim(\rvy)$ & \{128, 256, 512, 1024\} &  \\

Synthesis transform $g_s$  & Same as $g_a$ &  \\
Optimiser & \{Adam\} & \\
Learning rate & \{$1\cdot 10^{-4}$\} & \\
Batch size per device & $\{128\}$ & \\
Num devices & $\{1\}$ & \\

\bottomrule
\end{tabular}
\end{table}
\begin{table}[h]
\footnotesize
\centering
\caption{Hyperparameters for compression experiments on UCF-101.}
\label{tab:compress_ucf_hps}
\begin{tabular}{llc}
\toprule
\textbf{Parameter} & \textbf{Considered range} & \textbf{Comment} \\ 
\toprule
\textbf{INR training} & & \\
Patching & \{$(4, 8, 8), (8, 8, 8), (4, 16, 16), (8, 16, 16)$\} & Dividing $(24, 128, 128)$ dim. video.\\
Activation function & $\{h(x): \sin(\omega_0x) \text{ (SIREN)}\}$ & \\
$\omega_0$ & \{30\} & \\
Network depth & $\{10\}$ & \\
Network width & $\{256\}$ & \\

Batch size per device & $\{4\}$ & \\
Num devices & $\{4\}$ & \\
Optimiser & \{Adam\} & \\
Outer learning rate & \{$3\cdot 10^{-6}$\} & \\
Num inner steps & \{3\} & \\
Meta-learn $\boldsymbol\phi$ init. & \{True\} & \\
Meta SGD range & \{[-5.0, 5.0]\} & (Max./Min. for Meta-SGD LRs)\\
Meta SGD init range & \{[1.0, 1.0]\} & (Uniformly sampled).\\

\midrule
\textbf{$\boldsymbol\phi \rightarrow \{\rmG_{\mathtt{low}}^{(1)}, \dots, \rmG_{\mathtt{low}}^{(L)}\}$ network} & & \\
\midrule

$\dim(\boldsymbol\phi)$ & \{512, 1536, 2048, 2048\} &  \\
Use LayerNorm & \{True\} &  \\
Network width & \{512\} &  \\
Residual blocks & $\{2\}$ & \\
Activation function & \{Leaky Relu\} & \citep{xu2015empirical}\\
Adapt first Layer & \{False\} & Apply low-rank gating to 1st layer?\\

\midrule
\textbf{Quantiser training} & & \\
\midrule
Normalise $\boldsymbol\phi$ & $\{\text{True}\}$ & Per dim. $\frac{\boldsymbol\phi_i - \hat{\boldsymbol\mu}_i}{\hat{\boldsymbol\sigma}_i}$ based on $\boldsymbol\phi$ train-set stats.\\
$\lambda$ ($\mathcal{L}_{\mathtt{distortion}}$ penalty) & \{0.001, 0.01, 0.1, 1.0, 10.0\} & \\
Analysis transform ($g_a$) Residual blocks & $\{1\}$ & \\
$g_a$ Network width & $\{256, 512, 1024, 2048\}$ & \\
$g_a$ Activation function & SeLU & \citep{klambauer2017self} \\
$\dim(\rvy)$ & \{256, 512, 1024, 2048\} &  \\

Synthesis transform $g_s$ & Same as $g_a$ &  \\
Optimiser & \{Adam\} & \\
Learning rate & \{$1\cdot 10^{-4}$\} & \\
Batch size per device & $\{64\}$ & \\
Num devices & $\{1\}$ & \\

\bottomrule
\end{tabular}
\end{table}

\end{document}